\documentclass{article}

\usepackage{microtype}
\usepackage{graphicx}
\usepackage{subcaption}
\usepackage{booktabs} 

\usepackage{hyperref}

\usepackage[accepted]{icml2024}

\usepackage{amsmath}
\usepackage{amssymb}
\usepackage{mathtools}
\usepackage{amsthm}
\usepackage{tikz}
\usetikzlibrary{graphs,graphs.standard,arrows}
\usepackage[capitalize,noabbrev]{cleveref}

\usepackage[textsize=tiny]{todonotes}

\newtheorem{example-set}{Example}
\newtheorem{theorem}{Theorem}
\newtheorem{definition}{Definition}

\newtheorem{lemma}{Lemma}

\newtheorem{problem definition}{Problem Definition}

\newtheorem{remark}{Remark}

\usepackage{enumitem}

\usepackage{float}

\renewcommand{\algorithmicrequire}{ \textbf{Input:}} 
\renewcommand{\algorithmicensure}{ \textbf{Output:}} 

\usepackage{threeparttable}

\usepackage{centernot}

\newcommand{\CI}{\mathrel{\perp\mspace{-10mu}\perp}}
\newcommand{\nCI}{\centernot{\CI}}

\newdimen\arrowsize
\pgfarrowsdeclare{arcsq}{arcsq}
{
	\arrowsize=0.2pt
	\advance\arrowsize by .5\pgflinewidth
	\pgfarrowsleftextend{-4\arrowsize-.5\pgflinewidth}
	\pgfarrowsrightextend{.5\pgflinewidth}
}
{
	\arrowsize=0.8pt
	\advance\arrowsize by .5\pgflinewidth
	\pgfsetdash{}{0pt} 
	\pgfsetroundjoin   
	\pgfsetroundcap    
	\pgfpathmoveto{\pgfpoint{0\arrowsize}{0\arrowsize}}
	\pgfpatharc{-90}{-140}{4\arrowsize}
	\pgfusepathqstroke
	\pgfpathmoveto{\pgfpointorigin}
	\pgfpatharc{90}{140}{4\arrowsize}
	\pgfusepathqstroke
}

\tikzset{
  double arrow/.style={
    arrows={arcsq-arcsq}
  }
}

\newenvironment{nospaceflalign*}
 {\setlength{\abovedisplayskip}{1pt}\setlength{\belowdisplayskip}{1pt}%
  \csname flalign*\endcsname}
 {\csname endflalign*\endcsname\ignorespacesafterend}
 
 \newenvironment{nospaceflalign}
 {\setlength{\abovedisplayskip}{1pt}\setlength{\belowdisplayskip}{1pt}%
  \csname flalign\endcsname}
 {\csname endflalign\endcsname\ignorespacesafterend}

\icmltitlerunning{Local Causal Structure Learning in the Presence of Latent Variables}

\begin{document}

\twocolumn[
\icmltitle{Local Causal Structure Learning in the Presence of Latent Variables}




\begin{icmlauthorlist}
\icmlauthor{Feng Xie}{aaa}
\icmlauthor{Zheng Li}{aaa}
\icmlauthor{Peng Wu}{aaa}
\icmlauthor{Yan Zeng}{aaa}
\icmlauthor{Chunchen Liu}{bbb}
\icmlauthor{Zhi Geng}{aaa}
\end{icmlauthorlist}

\icmlaffiliation{aaa}{Department of Applied Statistics, Beijing Technology and Business University, Beijing, China}
\icmlaffiliation{bbb}{LingYang Co.Ltd, Alibaba Group, Hangzhou, China}
\icmlcorrespondingauthor{Yan Zeng}{yanazeng013@btbu.edu.cn}

\icmlkeywords{Machine Learning, ICML}

\vskip 0.3in
]



\printAffiliationsAndNotice{} 


\begin{abstract}
Discovering causal relationships from observational data, particularly in the presence of latent variables, poses a challenging problem.  
While current local structure learning methods have proven effective and efficient when the focus lies solely on the local relationships of a target variable, they operate under the assumption of causal sufficiency.  
This assumption implies that all the common causes of the measured variables are observed, leaving no room for latent variables. Such a premise can be easily violated in various real-world applications, resulting in inaccurate structures that may adversely impact downstream tasks.
In light of this, our paper delves into the primary investigation of locally identifying potential parents and children of a target from observational data that may include latent variables. 
Specifically, we harness the causal information from m-separation and V-structures to derive theoretical consistency results, effectively bridging the gap between global and local structure learning. 
Together with the newly developed stop rules, 
we present a principled method for determining whether a variable is a direct cause or effect of a target.
Further, we theoretically demonstrate the correctness of our approach under the standard causal Markov and faithfulness conditions, with infinite samples.
Experimental results on both synthetic and real-world data validate the effectiveness and efficiency of our approach.
\end{abstract}

\section{Introduction}
Inferring causal relations, known as causal discovery, has drawn much attention in several fields, such as computer science \cite{peters2017elements,pearl2018theoretical,scholkopf2022causality}, social science \cite{spirtes2000causation}, epidemiology \cite{hernan2010causal}, biology \cite{glymour2019review}, and neuroscience \cite{smith2011network,sanchez2019estimating}. The discovered causal relationships are useful for predicting the behavior of a system under external interventions, which is a crucial step in both understanding and manipulating that system \cite{pearl2009causality}. Learning such relations from purely observational data is challenging, especially when latent confounders can be present \cite{spirtes2016causal}.

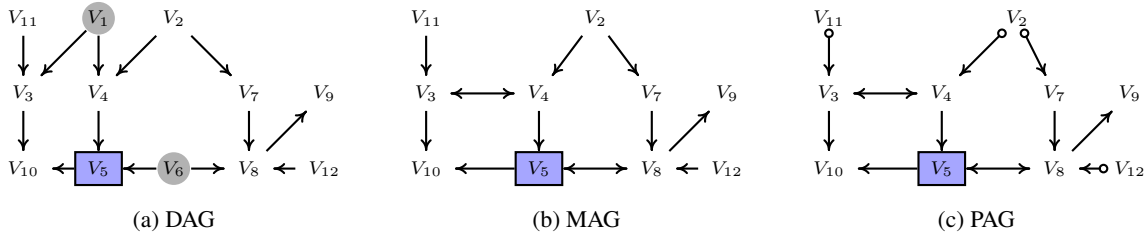
\begin{figure*}[htp]
    \centering
    \begin{tikzpicture}[node distance=1.5cm and 2cm, thick]
  \draw (0.0, 0.0) node(V10) [] {{\scriptsize\,$V_{10}$\,}};
  \draw (1.0, 0.0) node(V5) [rectangle, draw, fill=blue!35, minimum size=1mm] {{\scriptsize\,$V_{5}$\,}};
  \draw (2.0, 0.0) node(V6) [circle, fill=gray!60, inner sep=0pt, minimum size=3mm] {{\scriptsize\,$V_{6}$\,}};
  \draw (3.0, 0.0) node(V8) [] {{\scriptsize\,$V_{8}$\,}};
  \draw (4.0, 0.0) node(V12) [] {{\scriptsize\,$V_{12}$\,}};
  \draw (0.0, 1.0) node(V3) [] {{\scriptsize\,$V_{3}$\,}};
  \draw (1.0, 1.0) node(V4) [] {{\scriptsize\,$V_{4}$\,}};
  \draw (3.0, 1.0) node(V7) [] {{\scriptsize\,$V_{7}$\,}};
  \draw (4.0, 1.0) node(V9) [] {{\scriptsize\,$V_{9}$\,}};
  \draw (0.0, 2.0) node(V11) [] {{\scriptsize\,$V_{11}$\,}};
  \draw (1.0, 2.0) node(V1) [circle, fill=gray!60, inner sep=0pt, minimum size=3mm] {{\scriptsize\,$V_{1}$\,}};
  \draw (2.0, 2.0) node(V2) [] {{\scriptsize\,$V_{2}$\,}};
  \draw[-arcsq] (V11) -- (V3) node[pos=0.5,sloped,above] {};
  \draw[-arcsq] (V1) -- (V3) node[pos=0.5,sloped,above] {};
  \draw[-arcsq] (V1) -- (V4) node[pos=0.5,sloped,above] {};
  \draw[-arcsq] (V2) -- (V4) node[pos=0.5,sloped,above] {};
  \draw[-arcsq] (V2) -- (V7) node[pos=0.5,sloped,above] {};
  \draw[-arcsq] (V3) -- (V10) node[pos=0.5,sloped,above] {};
  \draw[-arcsq] (V4) -- (V5) node[pos=0.5,sloped,above] {};
  \draw[-arcsq] (V7) -- (V8) node[pos=0.5,sloped,above] {};
  \draw[-arcsq] (V5) -- (V10) node[pos=0.5,sloped,above] {};
  \draw[-arcsq] (V6) -- (V5) node[pos=0.5,sloped,above] {};
  \draw[-arcsq] (V6) -- (V8) node[pos=0.5,sloped,above] {};
  \draw[-arcsq] (V12) -- (V8) node[pos=0.5,sloped,above] {};
  \draw[-arcsq] (V8) -- (V9) node[pos=0.5,sloped,above] {};
  %
  \draw (2.0, -0.7) node(con3) [] {{\small\,(a) DAG\,}};
    \end{tikzpicture}~~~~~~
    \begin{tikzpicture}[node distance=1.5cm and 2cm, thick]
  \draw (0.0, 0.0) node(V10) [] {{\scriptsize\,$V_{10}$\,}};
  \draw (1.5, 0.0) node(V5) [rectangle, draw, fill=blue!35, minimum size=1mm] {{\scriptsize\,$V_{5}$\,}};
  \draw (3.0, 0.0) node(V8) [] {{\scriptsize\,$V_{8}$\,}};
  \draw (4.0, 0.0) node(V12) [] {{\scriptsize\,$V_{12}$\,}};
  \draw (0.0, 1.0) node(V3) [] {{\scriptsize\,$V_{3}$\,}};
  \draw (1.5, 1.0) node(V4) [] {{\scriptsize\,$V_{4}$\,}};
  \draw (3.0, 1.0) node(V7) [] {{\scriptsize\,$V_{7}$\,}};
  \draw (4.0, 1.0) node(V9) [] {{\scriptsize\,$V_{9}$\,}};
  \draw (0.0, 2.0) node(V11) [] {{\scriptsize\,$V_{11}$\,}};
  \draw (2.25, 2.0) node(V2) [] {{\scriptsize\,$V_{2}$\,}};
  \draw[-arcsq] (V11) -- (V3) node[pos=0.5,sloped,above] {};
  \draw[double arrow] (V3) -- (V4)
   node[pos=0.5,sloped,above] {};
  \draw[-arcsq] (V2) -- (V4) node[pos=0.5,sloped,above] {};
  \draw[-arcsq] (V2) -- (V7) node[pos=0.5,sloped,above] {};
  \draw[-arcsq] (V3) -- (V10) node[pos=0.5,sloped,above] {};
  \draw[-arcsq] (V4) -- (V5) node[pos=0.5,sloped,above] {};
  \draw[-arcsq] (V7) -- (V8) node[pos=0.5,sloped,above] {};
  \draw[-arcsq] (V5) -- (V10) node[pos=0.5,sloped,above] {};
  \draw[double arrow] (V5) -- (V8)
   node[pos=0.5,sloped,above] {};
  \draw[-arcsq] (V12) -- (V8) node[pos=0.5,sloped,above] {};
  \draw[-arcsq] (V8) -- (V9) node[pos=0.5,sloped,above] {};
  %
  \draw (2.0, -0.7) node(con3) [] {{\small\,(b) MAG\,}};
    \end{tikzpicture}~~~~~~
    \begin{tikzpicture}[node distance=1.5cm and 2cm, thick]
  \draw (0.0, 0.0) node(V10) [] {{\scriptsize\,$V_{10}$\,}};
  \draw (1.5, 0.0) node(V5) [rectangle, draw, fill=blue!35, minimum size=1mm] {{\scriptsize\,$V_{5}$\,}};
  \draw (3.0, 0.0) node(V8) [] {{\scriptsize\,$V_{8}$\,}};
  \draw (4.0, 0.0) node(V12) [] {{\scriptsize\,$V_{12}$\,}};
  \draw (0.0, 1.0) node(V3) [] {{\scriptsize\,$V_{3}$\,}};
  \draw (1.5, 1.0) node(V4) [] {{\scriptsize\,$V_{4}$\,}};
  \draw (3.0, 1.0) node(V7) [] {{\scriptsize\,$V_{7}$\,}};
  \draw (4.0, 1.0) node(V9) [] {{\scriptsize\,$V_{9}$\,}};
  \draw (0.0, 2.0) node(V11) [] {{\scriptsize\,$V_{11}$\,}};
  \draw (2.5, 2.0) node(V2) [] {{\scriptsize\,$V_{2}$\,}};
  \draw[-arcsq] (V11) -- (V3) node[pos=0.5,sloped,above] {};
  \draw (0.0, 1.80) circle (0.05);
  \draw[double arrow] (V3) -- (V4) node[pos=0.5,sloped,above] {};
  
  \draw[-arcsq] (V2) -- (V4) node[pos=0.5,sloped,above] {};
  \draw (2.3, 1.8) circle (0.05);
  
  \draw[-arcsq] (V2) -- (V7) node[pos=0.5,sloped,above] {};
  \draw (2.6, 1.8) circle (0.05);
  \draw[-arcsq] (V3) -- (V10) node[pos=0.5,sloped,above] {};
  
  \draw[-arcsq] (V4) -- (V5) node[pos=0.5,sloped,above] {};
  \draw[-arcsq] (V7) -- (V8) node[pos=0.5,sloped,above] {};
  \draw[-arcsq] (V5) -- (V10) node[pos=0.5,sloped,above] {};
  \draw[double arrow] (V5) -- (V8)
   node[pos=0.5,sloped,above] {};
  \draw[-arcsq] (V12) -- (V8) node[pos=0.5,sloped,above] {};
  \draw (3.65, 0.0) circle (0.05);
  \draw[-arcsq] (V8) -- (V9) node[pos=0.5,sloped,above] {};
  %
  \draw (2.0, -0.7) node(con3) [] {{\small\,(c)  PAG \,}};
    \end{tikzpicture}
    \vspace{-2mm}
    \caption{(a) Underlying causal DAG from a selected part of ANDES network \cite{conati1997line}, where $V_1$ and $V_6$ are hidden and $V_5$ is the target variable of interest. (b) The corresponding MAG of the DAG in (a). (c) The inferred PAG  from observed variables .}
    \label{fig-main-example}
\end{figure*}

There exists work in the literature that has attempted to recover causal structure among observed variables in the presence of latent variables. \citet{spirtes2000causation} proposed the seminal FCI (Fast Causal Inference) algorithm that can learn a partial ancestral graph (PAG) \footnote{A PAG represents a Markov equivalence class of maximal ancestral graphs (MAGs) which encode the causal relations among the observed variables. See the example in \autoref{fig-main-example} or Section \ref{Subsection-graph-notations} for more details.} in the presence of latent variables by performing conditional independence tests. Later, a faster algorithm, called Really Fast Causal Inference (RFCI), was developed \cite{colombo2012learning}. Other interesting developments along this line include \cite{claassen2011logical,claassen2013learning,raghu2018comparison,akbari2021recursive,mokhtarian2023novel}.
These works focus on learning the whole causal graph rather than the local causal graph. However, in many real-world scenarios, researchers are usually interested in the local causal relationships~\cite{walters2007genome,peter2011gene,ma2023local}.

Several contributions have been made to learn the local causal structure other than the global causal structure. For instance, the Local Causal Discovery (LCD) algorithm \cite{cooper1997simple} and its variants~\cite{silverstein2000scalable,mani2004causal} are proposed to find causal edges among every four-variable set in a causal graph. Although these algorithms primarily aim to identify a subset of causal edges through specific structures among all variables, our focus is on discovering all causal edges adjacent to a single target variable.
\citet{yin2008partial} and \citet{zhou2010discover} designed the PCD-by-PCD to find sets of parents, children, and maybe some of the descendants (PCD) of variables of the target variable.
Later, \citet{wang2014discovering} proposed a more efficient approach, called MB-by-MB, for discovering direct cause and effect variables of the target. Additional significant contributions to this field have been made, including the Causal Markov Blanket (CMB) algorithm \cite{gao2015local}, the Efficient Local Causal Structure (ELCS) algorithm \cite{yang2021towards}, and the GradieNt-based LCS (GraN-LCS) algorithm \cite{liang2023gradient}. Although these methods have been used in a range of fields, they usually assume the assumption of causal sufficiency, i.e., we have measured all the common causes of the measured variables in the system.
However, in various real-world scenarios, including 
Gene Expression network~\cite{wille2004sparse}, 
, etc, 
this assumption is usually violated.

In this paper, we address the challenge of locally learning the causes and effects of a given target variable in a more complex scenario where certain variables may be unmeasured. Specifically, our primary contributions can be summarized in three key aspects:
\begin{itemize}[leftmargin=15pt,itemsep=0pt,topsep=0pt,parsep=0pt]
    \item [1.] We propose a novel MMB-by-MMB algorithm for learning the direct causes and effects of a target variable based only on the estimated local structure, allowing for the existence of latent variables.
    \item [2.] We theoretically demonstrate that the proposed algorithm is a complete local discovery algorithm and can identify the same direct causes and effects for a target variable as global methods under standard assumptions.
    \item [3.] We conduct extensive experiments and demonstrate the efficacy of our algorithm on both benchmark network structures and real-world data.
\end{itemize}

\section{Related Works}
This paper focuses on local causal structure (LCS) learning. Our investigation intersects with broader themes, such as global causal structure (GCS) learning and Markov Blanket (MB) learning. In this context, we here provide a brief review of these three interconnected areas. For a comprehensive review of causal structure learning or MB learning, see   \cite{spirtes2016causal,heinze2018causal,yu2020causality,kitson2023survey}

\textbf{{LCS learning.}} Existing LCS learning methods can be roughly divided into two categories, namely Y-structure-based methods including LCD algorithm \cite{cooper1997simple} and its variants~\cite{silverstein2000scalable,mani2004causal,versteeg2022local}, and constraint-based ones such as  PCD-by-PCD \cite{yin2008partial,zhou2010discover}, MB-by-MB \cite{wang2014discovering}, CMB \cite{gao2015local}, ELCS \cite{yang2021towards}, and GraN-LCS algorithm \cite{liang2023gradient}. Methods in the first category typically focus on learning causal edges among sets of four variables, while our approach targets all causal edges related to a specific variable. Moreover, methods in the second category generally assume that all common causes of the measured variables are observed, an assumption not required by our method.

\textbf{{GCS learning.}} When latent confounding is present, well-known algorithms along this line include
the seminal FCI algorithm \cite{spirtes2000causation}, RFCI \cite{colombo2012learning}, FCI$^{+}$ \cite{claassen2013learning}, and its variants \cite{claassen2011logical,ogarrio2016hybrid,raghu2018comparison,akbari2021recursive}.
Some further studies are also conducted by introducing the data-generating mechanism \cite{chen2023discovery,kaltenpoth2023nonlinear,chen2021integer} or distribution of data \cite{hoyer2008estimation,salehkaleybar2020learning,maeda2020rcd,cai2023}. While these algorithms are efficient in their operation, identifying the global structure can be unnecessary and wasteful when our primary interest lies in understanding the local structure surrounding a single target variable, which can be clearly observed in the $nTest$ in our experimental results.

\textbf{{MB learning.}} MB learning algorithm aims to learn parents, children, and spouses of target T simultaneously. Along this line include GSMB \cite{margaritis1999bayesian}, IAMB \cite{tsamardinos2003towards}, Fast-IAMB \cite{yaramakala2005speculative}, and Total
Conditioning (TC)\cite{pellet2008using}, and other variants \cite{aliferis2003hiton,pena2007towards,gao2016efficient,wu2019accurate}. Recently, \citet{yu2018mining} proposed an algorithm, M3B to mine the MAG MB (MMB) of a target variable in MAGs.
However, the above methods do not distinguish parents from children.
In contrast, our method has to differentiate the direct parents (cause) and children (effect). 

To the best of our knowledge, there is currently no method for learning the local causal structure in the presence of latent confounders that can effectively identify the direct causes and effects of a target variable under standard assumptions. 

\section{Preliminaries}
\subsection{Graph Terminology and Notations}\label{Subsection-graph-notations}
\textbf{Ancestral Graphs.} A \emph{\textbf{mixed graph}} $\mathcal{G}$ over the set of vertices $\mathbf{V}$ containing three types of edges between pairs of nodes: directed edges ($\to$), bi-directed edges ($\leftrightarrow$), and undirected edges ($\--$). A is a \emph{\textbf{spouse}} of B if $A \leftrightarrow B$ is in $\mathcal{G}$. A mixed graph is \emph{\textbf{ancestral}} if it doesn’t contain a directed or almost directed
cycle \footnote{An almost directed cycle happens when A is both a spouse and an ancestor of B.}. Let $\mathcal{V}$ be any subset of vertices in $\mathcal{G}$. An \emph{\textbf{inducing path}} relative to $\mathcal{V}$ is a path on which every vertex not in $\mathcal{V}$ (except for the endpoints) is a collider on the path and every collider is an ancestor of an endpoint of the path.
An ancestral graph is a \emph{\textbf{Maximal Ancestral Graph (MAG)}} $\mathcal{M}$ if there is no inducing path between any two non-adjacent vertices.
A MAG is called a directed acyclic graph (DAG) if it has only directed edges. A causal MAG represents a set of causal models with the same set of observed variables that entail the same independence and ancestral relations among the observed variables.
Two MAGs are called \emph{\textbf{Markov equivalent}} if they impose the same independence model. A \emph{\textbf{Partial Ancestral Graph} (PAG)}$\mathcal{P}$ represents an equivalence class of MAGs $[\mathcal{M}]$. A partial ancestral graph for $[\mathcal{M}]$ is a graph $\mathcal{P}$ with possible three kinds of marks ($\circ$, $>$, $\--$)\footnote{$\circ$ represents undetermined edge marks.}, such that 1) $\mathcal{P}$ has the same adjacencies as $\mathcal{M}$ (and hence any member of $[\mathcal{M}]$) does, and every non-circle mark in $\mathcal{P}$ is an invariant mark in $[\mathcal{M}]$. For convenience, we use an asterisk (*) to denote any possible mark of a PAG ($\circ,>,\--$) or a MAG ($>,\--$).

\begin{definition}[m-separation]
    In a mixed graph $\mathcal{G}$, a path ${p}$ between vertices $X$ and $Y$ is active (m-connecting) relative to a (possibly empty) set of vertices $\mathbf{Z}$ ($X, Y \notin \mathbf{Z}$) if 1) every non-collider on $p$ is not a member of $\mathbf{Z}$, and 2) every collider on $p$ has a descendant in $\mathbf{Z}$.
\end{definition}
A set $\mathbf{Z}$ m-separates $\mathbf{X}$ and $\mathbf{Y}$ in $\mathcal{G}$, denoted by $(\mathbf{X} \CI \mathbf{Y} | \mathbf{Z})_{\mathcal{G}}$, if there is no active path between any vertices in $\mathbf{X}$ and any vertices in $\mathbf{Y}$ relative to $\mathbf{Z}$. The criterion of m-separation is a generalization of Pearl's d-separation criterion in DAG to ancestral graphs.
Two MAGs are called Markov equivalent if they impose the same m-separations.

Related concepts used here can be found in sources \cite{richardson2002ancestral,zhang2008completeness}.

\textbf{Markov Blanket.} In a DAG, the \emph{\textbf{Markov blanket}} of a vertice $T$, noted $\mathit{MB(T)}$, is the set of parents, children, and children’s parents (spouses) of $T$. In a MAG, the Markov blanket of a vertice $T$, noted as $\mathit{MMB(T)}$, consists of 1) parents of $T$; 2) children of $T$; and 3) a set of variables that for $\forall V_i$ within the set, $V_i$ is not adjacent to $T$ and has a collider path to $T$. See the example in the Definition \autoref{definition-MMB}. 

\textbf{Notations.} Given a graph $\mathcal{G}$, two vertices are said to be adjacent in $\mathcal{G}$ if there is an edge between them. We use $\mathit{Adj(T)}$ to denote the set of adjacent vertices of $T$.
$X$ is called an ancestor of $Y$ and $Y$ a descendant
of $X$ if there is a directed path from $X$ to $Y$ or $X=Y$. 
We use $\mathit{Pa(T)}$, $\mathit{Ch(T)}$, $\mathit{Sp(T)}$, $\mathit{An(T)}$, $\mathit{De(T)}$ to denote the set of \emph{\textbf{parents, children, spouses, ancestors}}, and \emph{\textbf{descendants}} of vertex $T$ in $\mathcal{G}$, respectively. 
We use the notation $(\mathbf{X} \CI \mathbf{Y} | \mathbf{Z})_{P}$ for ``$\mathbf{X}$ is statistically independent of $\mathbf{Y}$ given $\mathbf{Z}$”, and $(\mathbf{X} \nCI \mathbf{Y} | \mathbf{Z})_{P}$ for the negation of the same sentence \citep{dawid1979conditional}. We drop the subscript $P$ whenever it is clear from context. We use $\mathit{MMB^{+}(T)}$ to denote the set of $\{\mathit{MMB(T)} \cup T\}$. 

\textbf{Standard Assumption.}  In terms of m-separation, the \emph{\textbf{causal
Markov condition}} says that m-separation in a graph $\mathcal{G}$ implies conditional independence in the population distribution. The \emph{\textbf{causal Faithfulness condition}} says that m-connection in a graph $\mathcal{G}$ implies conditional dependence in the population distribution \citep{zhang2008causal}. 
Under the above two conditions, conditional independence relations among the observed variables correspond exactly to m-separation in the MAG $\mathcal{G}$, i.e., $(\mathbf{X} \CI \mathbf{Y} | \mathbf{Z})_{P} \Leftrightarrow (\mathbf{X} \CI \mathbf{Y} | \mathbf{Z})_{\mathcal{G}}$.

\textbf{Identification of Global Learning for PAG.}  Assuming the causal Markov condition and the causal Faithfulness condition, the PAG (that represents an equivalence class of MAGs) can be uniquely identified by using the independence-constraint-based algorithm, such as FCI \citep{spirtes2000causation,zhang2008completeness}, from an oracle of conditional independence relations.

\subsection{Problem Definition}
We consider a Structural Causal Model (SCM) \citep{pearl2009causality} with the set of variables $\mathbf{V} =\mathbf{O} \cup \mathbf{L}$, and the joint distribution $P({\mathbf{V})}$, where $\mathbf{O}$ and $\mathbf{L}$ denote the set of observed variables and latent variables, respectively. We here assume that there is no selection bias in the system. Thus, the SCM is associated with a DAG where each node is a variable in $\mathbf{V}$ and each edge is a function $f$. That is to say, each variable $V_i \in \mathbf{V}$ is generated as $V_i = f_i(Pa(V_i), u_i)$, where $u_i$ represent errors (or “disturbances”) due to omitted factors, and all errors are independent from each other.

\textbf{Goal.} Given a target variable $T \in \mathbf{O}$,  we are interested in the local structure of the target variable. In particular, our goal is to establish the local criteria for identifying the potential direct causes and effects of a target based only on the local structure instead of the entire graph.

\section{MMB-by-MMB Algorithm}
In this section, we propose a sequential algorithm, MMB-by-MMB, for discovering the direct causes and effects of a target variable $T$.
We use $\mathbf{Waitlist}$ to store nodes that are potentially relevant for identifying the direct causes and effects of $T$. Let $\mathbf{Donelist}$ store nodes removed from $\mathbf{Waitlist}$ and $\mathcal{P}$ store valid causal information.

\textbf{Basic idea.} 
The process proceeds through a series of sequential repeats. Initially, $\mathbf{Waitlist}=\{T\}$ and $\mathbf{Donlist}$ is empty. In each iteration, we focus on the first variable $V_X$ in the $\mathbf{Waitlist}$. 
Specifically, we first learn the MMB of $V_X$ and a local causal structure $\mathcal{L}_{\mathit{MMB}^+(V_X)}$ based on this MMB. 
Subsequently, employing m-separation (see Theorem \ref{theorem1}) and V-structure (see Theorem \ref{theorem2}), we select true causal information from $\mathcal{L}_{\mathit{MMB}^+(V_X)}$ and store them in $\mathcal{P}$. 
Next, $\mathcal{P}$ is oriented using standard orientation criteria. Finally, we update $\mathbf{Waitlist}$ and $\mathbf{Donlist}$. When the stop rules are met (see Theorem \ref{theorem3}), the algorithm stops running. The specific pseudocode is provided in Algorithm \ref{MMB-by-MMB} and \ref{MMB-alg}.

We outline the principle of the algorithm in Section \ref{Subsec-Principle}. Furthermore, we present the detailed steps of the algorithm in Section \ref{Subsec-Algorithm}. Under standard assumptions, we show that the proposed algorithm can locally identify the same direct causes and effects for a target variable as global learning methods.
Finally, in Section \ref{Subsec-Complexity}, we analyze the complexity of the algorithm. To improve readability, we defer all proofs to Appendix \ref{Appendix-proofs}.

\subsection{Principle of the Algorithm}\label{Subsec-Principle}
In this section, we present the theoretical results that serve as the principle for our sequential approach.
Specifically, we answer the following 3 questions:
\begin{description}
[leftmargin=14pt,itemsep=0pt,topsep=0pt,parsep=0pt]
\item [$\mathcal{Q}1$.] What causal information of \textbf{\emph{m-separation}} in local structure learning is consistent with those in global learning?
\item [$\mathcal{Q}2$.] What causal information of \textbf{\emph{V-structures}} in local structure learning is consistent with those in global learning?
\item [$\mathcal{Q}3$.] How to design a \textbf{\emph{stop}} criteria to ensure that our local learning structure is consistent with the global one?
\end{description}
We first give the following theorem about m-separation in both local and global learning, which answers question $\mathcal{Q}1$.
\begin{theorem}[\textbf{M-separation}]\label{theorem1}
Let $T$ be any node in $\mathbf{O}$, and $X$ be a node in $\mathit{MMB(T)}$. Then $T$ and $X$ are m-separated by a subset of $\mathbf{O} \setminus \left \{ T, X \right \}$ if and only if they are m-separated by a subset of $\mathit{MMB(T)}\setminus \left \{ X \right \}$.
\end{theorem}
Theorem \ref{theorem1} implies that the existence of an edge connecting $T$ to any other node $X\in \mathit{MMB(T)}$ can be equivalently determined through both the full distribution of $\mathbf{O}$ and the marginal distribution of $\mathit {MMB^{+}(T)}$. Consequently, it becomes feasible to accurately assess the presence of these connecting edges to $T$ by utilizing the observed data from $\mathit {MMB^{+}(T)}$.

\begin{example-set}
Consider the MAG shown in \autoref{fig-main-example}(b). 
Let $V_5$ be the target variable $T$. Suppose that we can correctly check conditional independencies from data and thus find the $\mathit{MMB(V_5)}$, i.e., $\mathit{MMB(V_5)}=\{V_3, V_4, V_7, V_8, V_{10}, V_{12}\}$. According to \autoref{theorem1}, we deduce the existence of edges $V_5 \circ\!\!\--\!\!\circ V_4$, $V_5 \circ\!\!\--\!\!\circ V_8$ and $V_5 \circ\!\!\--\!\!\circ V_{10} \ $, while there are no connecting edges between $V_5$ and $V_3$, $V_7$, or $V_{12}$. These results are consistent with the conclusions of global learning.
\end{example-set}

\begin{remark}\label{remark:1}
It is noteworthy that the connecting edges between nodes in  $\mathit{MMB(T)}$ through the marginal distribution of $\mathit {MMB^{+}(T)}$ do not align with those identified through the full distribution of $\mathbf{O}$. For instance, considering the connection between $V_4$ and $V_7$, we will obtain the spurious edge $V_4 \circ\!\!\--\!\!\circ V_7$ from $\mathit {MMB^{+}(T)}$. However, because $V_4 \CI V_7|\ V_2$, we know there is no direct edge between $V_4$ and $ V_7$ in the ground-truth MAG.
\end{remark}

Next, we discuss the solution for the question $\mathcal{Q}2$, and the illustrative examples are given accordingly. Let $\mathcal{V}$ be a subset of $\mathbf{V}$. We say that a V-structure $X \rightarrow Z \leftarrow Y$ can be identified or found by the marginal distribution $P(\mathcal{V})$ if the conditional independence and dependence of the V-structure can be checked in the $P(\mathcal{V})$, i.e., $X \CI Y | \mathbf{S}$ and $X \nCI Y | \mathbf{S} \cup \{Z\}$ for $\{X,Y,Z\}\cup \mathbf{S} \subseteq \mathcal{V}$.

\begin{theorem}[\textbf{Fully Correct V-structures}]
Consider a sub-MAG of $\mathcal{M}^{\prime}$ over $\mathit{MMB^{+}(T)}$. Let $V_a, V_b$ be two nodes in $\mathit{MMB(T)}$. The following statements hold.
\begin{description}[leftmargin=14pt,itemsep=0pt,topsep=0pt,parsep=0pt]
    \item [$\mathcal{S}1$.] The V-structure $V_a *\!\!\rightarrow T \leftarrow\!\! * V_b$ that identified by the marginal distribution of $\mathit {MMB^{+}(T)}$ are true V-structures in the ground-truth MAG $\mathcal{M}$.
    \item [$\mathcal{S}2$.] The V-structure $T *\!\!\rightarrow V_a \leftarrow\!\! * V_b$ can be successfully identified by the marginal distribution of $\mathit {MMB^{+}(T)}$.
\end{description}
\label{theorem2}
\end{theorem}
Statement $\mathcal{S}1$ shows that if $T$ is a collider in the identified V-structures using the observational data of $\mathit{MMB^{+}(T)}$, then these V-structures are equivalent determined by the full observational data of $\mathbf{O}$.
Statement $\mathcal{S}2$ says that a special type of V-structure, in which the collider $V_a$ within the V-structure is not an ancestor of $T$, can certainly be identified from the observational data of $\mathit{MMB^{+}(T)}$. 

\begin{example-set}[Statements $\mathcal{S}1$ and $\mathcal{S}2$]
Continue to consider the causal diagram shown in \autoref{fig-main-example}(b). We have known $\mathit{MMB^{+}(V_5)}=\{V_5,V_3,V_4,V_7,V_8,V_{10},V_{12}\}$. According to Statement $\mathcal{S}1$, we can determine the V-structure $V_4 *\!\!\rightarrow V_5 \leftarrow\!\! * V_8$ form the marginal distribution of $\mathit{MMB^{+}(V_5)}$, since $V_4 \CI V_8 |V_7$ and $V_4 \nCI V_8 |\{V_7,V_5\}$. Furthermore, according to Statement $\mathcal{S}2$, we can obtain the V-structure $V_5 *\!\!\rightarrow V_8 \leftarrow\!\! * V_7$ from the marginal distribution of $\mathit {MMB^{+}(V_5)}$, since $\ V_5 \CI V_7|V_4 $ and $\ V_5 \nCI V_7|\{V_4,V_8\}$.
\end{example-set}
\begin{remark}
\autoref{theorem2} merely states that the locally identified V-structures containing T are correct. That is to say, during the orientation of local structures, some V-structures may be incorrect or missing, as stated in the following two examples. 
\begin{description}
[leftmargin=15pt,itemsep=0pt,topsep=0pt,parsep=0pt]
    \item [1.] V-structures identified that do not include $T$ cannot be guaranteed to be correct from the observational data of $\mathit{MMB^{+}(T)}$. For instance, consider the graph shown in \autoref{fig-main-example}.(b), one may obtain $V_3 *\!\!\rightarrow V_4 \leftarrow\!\! * V_7$ from the observational data of $\mathit{MMB^{+}(V_5)}$ since $\ V_3 \CI V_7|\ \emptyset $ and $\ V_3 \nCI V_7|\ V_4$. However, the V-structure $V_3 *\!\!\rightarrow V_4 \leftarrow\!\! * V_7$ is not entirely accurate. In global structural learning, the identified V-structure is $V_3 *\!\!\rightarrow V_4 \leftarrow\!\! * V_2$.
    \item [2.] Not all V-structures that include collider $T$ can be identified from the observational data of $\mathit{MMB^{+}(T)}$. For instance, let $V_{10}$ be the target variable in \autoref{fig-main-example}(b). Then we have $MMB(V_{10}) = \{ V_3, V_5\}$. The V-structure $V_3 *\!\!\rightarrow V_{10} \leftarrow\!\! * V_5$ is unidentifiable within $\mathit{MMB^{+}(V_{10})}$, as the separating set for $V_3$ and $V_5$ (i.e., $V_4$) is not encompassed in $\mathit{MMB(V_{10})}$.
\end{description}
\label{remark:2}
\end{remark}

We now address the last question $\mathcal{Q}3$ about the criteria for stopping rules.
\begin{theorem}[\textbf{Stop Rules}]
Let $T$ be the target node of interest within $\mathbf{O}$ and $\mathbf{Waitlist}$ represent the collection of nodes that need to be checked by \autoref{theorem1} and \autoref{theorem2}. If any of the subsequent rules are met, the local structure identified for $T$, encompassing its direct causes and effects, will be equivalent to the structure identified through global learning methods.
\begin{description}
[leftmargin=14pt,itemsep=0pt,topsep=3pt,parsep=0pt]
    \item $\mathcal{R}1.$ The structures around the target $T$ are all determined.
    \item $\mathcal{R}2.$ The $\mathbf{Waitlist}$ is empty.
    \item $\mathcal{R}3.$ All paths from the target $T$, which include undirected edges (connected to the target $T$), are blocked by edges $*\!\!\rightarrow$. 
\end{description}
\label{theorem3}
\end{theorem}
Rules $\mathcal{R}1$ and $\mathcal{R}2$ are both direct stopping criteria, meaning that all causal information of interest has been identified, or all nodes have been fully utilized.
Roughly speaking, R3 states that when the paths connecting the surrounding undirected edges of $T$ are blocked by directed edges ($*\!\!\rightarrow$), integrating the joint distribution of the remaining nodes connected on these paths is equivalent to omitting these nodes (see Lemma \ref{Lemma-Joint-probability} in the appendix).
In other words, continuing to learn the local structure of the remaining nodes connected on these paths will not help determine the direction of the undirected edges around $T$. 

Below, we give an example to illustrate the rule $\mathcal{R}3$ in \autoref{theorem3}.
\begin{figure}[htp!]
    \centering
    \begin{tikzpicture}[node distance=1.5cm and 2cm, thick]
    \draw (0.0, 0.0) node(T) [rectangle, draw, fill=blue!35, minimum size=1mm] {{\scriptsize\,$T$\,}};
    \draw (1.3, 0.0) node(V1) [] {{\scriptsize\,$V_{1}$\,}};
    \draw (1.3, 1.0) node(V2) [] {{\scriptsize\,$V_{2}$\,}};
    \draw (2.6, 1.0) node(V4) [] {{\scriptsize\,$V_{4}$\,}};
    \draw (2.6, 0.0) node(V3) [] {{\scriptsize\,$V_{3}$\,}};

    \draw[double arrow] (T) -- (V1) node[pos=0.5,sloped,above] {};
    \draw[-arcsq] (V2) -- (V1) node[pos=0.5,sloped,above] {};
    \draw[-arcsq] (V2) -- (V1) node[pos=0.5,sloped,above] {};
    \draw[-arcsq] (V1) -- (V3) node[pos=0.5,sloped,above] {};
    \draw[-arcsq] (V3) -- (V4) node[pos=0.5,sloped,above] {};
    
    \draw (1.3, -0.5) node(con3) [] {{\small\,(a) Underlying MAG\,}};
    \end{tikzpicture}~~
    \begin{tikzpicture}[node distance=1.5cm and 2cm, thick]
    \draw (0.0, 0.0) node(T) [] {{\scriptsize\,$T$\,}};
    \draw (1.3, 0.0) node(V1) [] {{\scriptsize\,$V_{1}$\,}};
    \draw (1.3, 1.0) node(V2) [] {{\scriptsize\,$V_{2}$\,}};

    \draw[-arcsq, color=blue, line width=1.5pt] (T) -- (V1) node[pos=0.5,sloped,above] {};
    \draw (0.2, 0.0) circle (0.05);
    \draw[-arcsq] (V2) -- (V1) node[pos=0.5,sloped,above] {};
    \draw (1.3, 0.8) circle (0.05);
    \draw (0.65, -0.5) node(con3) [] {{\small\,(b) The final local PAG around $T$\,}};
    \end{tikzpicture}
    \vspace{-3mm}
    \caption{The illustrative example for $\mathcal{R}3$ in \autoref{theorem3}.}
    \label{Fig-theorem-R3}
\vspace{-3mm}
\end{figure}
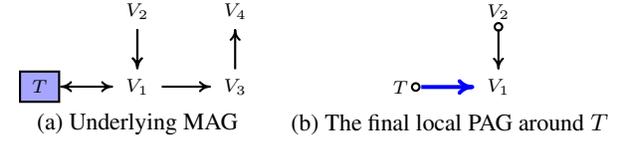
\begin{example-set}[$\mathcal{R}3$]
Consider the graph shown in \autoref{Fig-theorem-R3}(a), where $T$ is the target variable. Assuming that we have already checked nodes $T$ according to \autoref{theorem1} and \autoref{theorem2}, we then obtain subgraph (b). Note that the left tail of edge $T\circ \!\!\rightarrow V_1 $ is 
not directed. However, we can find that the path $T\circ \!\!\rightarrow V_1 $ is blocked by the edge $ \circ \!\! \rightarrow $. According to $\mathcal{R}3$ of \autoref{theorem3}, we will stop the learning process, even though nodes $V_3$ and $V_4$ have not yet been checked.

\end{example-set}
\begin{algorithm}[hpt!]
   \caption{MMB-by-MMB}
   \label{MMB-by-MMB}
   \begin{algorithmic}[1]
   \REQUIRE Target $T$, observed data $\mathbf{O}$ of $\mathbf{V}$
   \STATE Initialize : $\mathbf{Waitlist} \coloneqq \{T\}$, $\mathbf{Donelist} = \emptyset$,  $\mathcal{P} = \emptyset$.
   \REPEAT
     \STATE $V_X$ $\leftarrow$ the head node of $\mathbf{Waitlist}$;\\
     \STATE $\mathit{MMB^{+}(V_X)} \leftarrow \mathit{MMB_{alg}(V_X)}$ ;
     \IF{$\exists V_{Y}\!\in\!\mathbf{Donelist}, \mathit{MMB}^{+}(V_X)\subseteq \mathit{MMB^{+}(V_{Y})}$}
     \STATE $\mathcal{L}_{X}$ $\leftarrow$ the substructure of $\mathcal{L}_{Y}$ over $\mathit{MMB}^{+}(X)$;
     \ELSIF{$\mathit{MMB(V_X)} \subseteq \mathbf{Donelist}$}
     \STATE $\mathcal{L}_{X}$ $\leftarrow$ the substructure of $\mathcal{P}$ over $\mathit{MMB^{+}(X)}$;
     \ELSE
     \STATE Learn $\mathcal{L}_{X}$ over $\mathit{MMB^{+}(X)}$.
     \ENDIF
     \STATE $\mathcal{P}$ $\leftarrow$ select the edges connected to $V_X$ and the V-structures containing $V_X$.
     \STATE $\mathcal{P}$ $\leftarrow$ orient maximally the edge marks using the orientation rules of \citet{zhang2008completeness}.
     \STATE Add $V_X$ to $\mathbf{Donelist}$, and remove $V_X$ from the $\mathbf{Waitlist}$.
     \STATE Add $\{\mathit{Adj(V_X)} \setminus ( \mathbf{Waitlist} \cup \mathbf{Donelist})\}$ to $\mathbf{Waitlist}$
  \UNTIL{One of the stop Rules $\mathcal{R}1 \sim \mathcal{R}3$ is met} 
   \ENSURE The local structure $\mathcal{P}$ around $T$
   \end{algorithmic}
\end{algorithm}
\vspace{-2mm}
\begin{algorithm}[hpt!]
   \caption{$\mathit{MMB}_{alg}$\cite{pellet2008using}}
   \label{MMB-alg}
   \begin{algorithmic}[1]
   \REQUIRE Target $V_x$, observed data $\mathbf{O}$ of $\mathbf{V}$
   \STATE Initialize : $\mathit{MMB}(V_X) \coloneqq \emptyset$.
   \FOR{each $V_Y \in \mathbf{O} \setminus V_X$}
       \IF{$V_X \nCI V_Y \mid \mathbf{O} \setminus \{V_X, V_Y\}$}
           \STATE Add $V_Y$ to $\mathit{MMB}(V_X)$
       \ENDIF
   \ENDFOR
   \STATE $\mathit{MMB^{+}(V_{X})} = \{V_X \cup \mathit{MMB}(V_X)\}$
   \ENSURE $\mathit{MMB^{+}(V_{X})}$
   \end{algorithmic}
\end{algorithm}
\subsection{Our Sequential Approach}\label{Subsec-Algorithm}
\begin{figure*}[t]
    \centering
    \begin{subfigure}{0.3\textwidth}
    \begin{tikzpicture}[node distance=1.5cm and 2cm, thick]
    \draw (0.0, 0.0) node(V10) [] {{\scriptsize\,$V_{10}$\,}};
  \draw (1.5, 0.0) node(V5) [rectangle, draw, fill=blue!35, minimum size=1mm] {{\scriptsize\,$V_{5}$\,}};
  \draw (3.0, 0.0) node(V8) [] {{\scriptsize\,$V_{8}$\,}};
  \draw (4.0, 0.0) node(V12) [] {{\scriptsize\,$V_{12}$\,}};
  \draw (0.0, 1.0) node(V3) [] {{\scriptsize\,$V_{3}$\,}};
  \draw (1.5, 1.0) node(V4) [] {{\scriptsize\,$V_{4}$\,}};
  \draw (3.0, 1.0) node(V7) [] {{\scriptsize\,$V_{7}$\,}};
  \draw (4.0, 1.0) node(V9) [] {{\scriptsize\,$V_{9}$\,}};
  \draw (0.0, 2.0) node(V11) [] {{\scriptsize\,$V_{11}$\,}};
  \draw (2.25, 2.0) node(V2) [] {{\scriptsize\,$V_{2}$\,}};
  \draw[-arcsq] (V11) -- (V3) node[pos=0.5,sloped,above] {};
  \draw[double arrow] (V3) -- (V4)
   node[pos=0.5,sloped,above] {};
  \draw[-arcsq] (V2) -- (V4) node[pos=0.5,sloped,above] {};
  \draw[-arcsq] (V2) -- (V7) node[pos=0.5,sloped,above] {};
  \draw[-arcsq] (V3) -- (V10) node[pos=0.5,sloped,above] {};
  \draw[-arcsq] (V4) -- (V5) node[pos=0.5,sloped,above] {};
  \draw[-arcsq] (V7) -- (V8) node[pos=0.5,sloped,above] {};
  \draw[-arcsq] (V5) -- (V10) node[pos=0.5,sloped,above] {};
  \draw[double arrow] (V5) -- (V8)
   node[pos=0.5,sloped,above] {};
  \draw[-arcsq] (V12) -- (V8) node[pos=0.5,sloped,above] {};
  \draw[-arcsq] (V8) -- (V9) node[pos=0.5,sloped,above] {};
  %
    \draw (2.0, -0.7) node(con3) [] {{\small\,(a) Underlying MAG\,}};
\end{tikzpicture}~~~~~~
    \end{subfigure}
    \hspace{0.5em}
    \begin{subfigure}{0.3\textwidth}
    \begin{tikzpicture}[node distance=1.5cm and 2cm, thick]
      \draw [fill=red!50,thick, fill opacity=0.5, draw=none] (1.5,0.0) ellipse [x radius=0.5cm, y radius=0.35cm];
      \draw (0.0, 0.0) node(V10) [] {{\scriptsize\,$V_{10}$\,}};
      \draw (1.5, 0.0) node(V5) [] {{\scriptsize\,$V_{5}$\,}};
      \draw (3.0, 0.0) node(V8) [] {{\scriptsize\,$V_{8}$\,}};
      \draw (4.3, 0.0) node(V12) [] {{\scriptsize\,$V_{12}$\,}};
      \draw (0.0, 1.0) node(V3) [] {{\scriptsize\,$V_{3}$\,}};
      \draw (1.5, 1.0) node(V4) [] {{\scriptsize\,$V_{4}$\,}};
      \draw (3.0, 1.0) node(V7) [] {{\scriptsize\,$V_{7}$\,}};
      \draw[-arcsq] (V4) -- (V5) node[pos=0.5,sloped,above] {};
      \draw (1.5, 0.85) circle (0.05);
      \draw[-arcsq] (V5)  --  (V10) node[pos=0.5,sloped,above] {};
      \draw (1.25, 0.0) circle (0.05);
      \draw[-arcsq] (V3) -- (V10) node[pos=0.5,sloped,above] {};
      \draw (0, 0.85) circle (0.05);
      \draw[double arrow] (V8) -- (V5)
       node[pos=0.5,sloped,above] {};
      \draw[-arcsq] (V7) -- (V8) node[pos=0.5,sloped,above] {};
      \draw (3.0, 0.85) circle (0.05);
      \draw[-arcsq] (V12) -- (V8) node[pos=0.5,sloped,above] {};
      \draw (3.95, 0.0) circle (0.05);
      \draw[red] (0.25, 1.0) circle (0.05);
      \draw[-arcsq, red] (V3) -- (V4) node[pos=0.5,sloped,above] {};
      %
      \draw[-arcsq,red] (V7) -- (V4) node[pos=0.5,sloped,above] {};
      \draw[red] (2.75, 1.0) circle (0.05);
      \draw (2.0, -0.7) node(con3) [] {{\small\,(b) $\mathcal{L}_{\mathit{MMB^{+}}(V_5)}$ \,}};
    \end{tikzpicture}
    \end{subfigure}
    \hspace{0.5em}
    \begin{subfigure}{0.3\textwidth}
    \begin{tikzpicture}[node distance=1.5cm and 2cm, thick]
      \draw (0.0, 0.0) node(V10) [] {{\scriptsize\,$V_{10}$\,}};
      \draw (1.5, 0.0) node(V5) [] {{\scriptsize\,$V_{5}$\,}};
      \draw (3.0, 0.0) node(V8) [] {{\scriptsize\,$V_{8}$\,}};
      \draw (4.3, 0.0) node(V12) [] {{\scriptsize\,$V_{12}$\,}};
      \draw (0.0, 1.0) node(V3) [] {{\scriptsize\,$V_{3}$\,}};
      \draw (1.5, 1.0) node(V4) [] {{\scriptsize\,$V_{4}$\,}};
      \draw (3.0, 1.0) node(V7) [] {{\scriptsize\,$V_{7}$\,}};
      \draw[-arcsq] (V4) -- (V5) node[pos=0.5,sloped,above] {};
      \draw (1.5, 0.85) circle (0.05);
      \draw[-arcsq] (V5)  --  (V10) node[pos=0.5,sloped,above] {};
      \draw[-arcsq] (V3) -- (V10) node[pos=0.5,sloped,above] {};
      \draw (0, 0.85) circle (0.05);
      \draw[double arrow] (V8) -- (V5)
       node[pos=0.5,sloped,above] {};
      \draw[-arcsq] (V7) -- (V8) node[pos=0.5,sloped,above] {};
      \draw (3.0, 0.85) circle (0.05);
      \draw[-arcsq] (V12) -- (V8) node[pos=0.5,sloped,above] {};
      \draw (3.95, 0.0) circle (0.05);
      \draw (2.0, -0.7) node(con3) [] {{\small\,(c) Updated $\mathcal{P}$ after learning $\mathcal{L}_{\mathit{MMB^{+}}(V_5)}$ \,}};
    \end{tikzpicture}
    \end{subfigure}
    \hspace{0.5em}
    \begin{subfigure}{0.3\textwidth}
    \begin{tikzpicture}[node distance=1.5cm and 2cm, thick]
    \draw [fill=red!50,thick, fill opacity=0.5, draw=none] (1.5,1.0) ellipse [x radius=0.5cm, y radius=0.35cm];
      \draw (0.0, 2.0) node(V11) [] {{\scriptsize\,$V_{11}$\,}};
      \draw (1.5, 0.0) node(V5) [] {{\scriptsize\,$V_{5}$\,}};
      \draw (3.0, 0.0) node(V8) [] {{\scriptsize\,$V_{8}$\,}};
      \draw (4.3, 0.0) node(V12) [] {{\scriptsize\,$V_{12}$\,}};
      \draw (0.0, 1.0) node(V3) [] {{\scriptsize\,$V_{3}$\,}};
      \draw (1.5, 1.0) node(V4) [] {{\scriptsize\,$V_{4}$\,}};
      \draw (3.0, 1.0) node(V7) [] {{\scriptsize\,$V_{7}$\,}};
      \draw (2.25, 2.0) node(V2) [] {{\scriptsize\,$V_{2}$\,}};
      \draw[double arrow] (V3) -- (V4);
      \draw[-arcsq] (V2) -- (V4) node[pos=0.5,sloped,above] {};
      \draw [color=red] (2.4, 1.8) circle (0.05);
      
      \draw[-arcsq] (V4) -- (V5) node[pos=0.5,sloped,above] {};
      \draw (1.5, 0.85) circle (0.05);
      \draw[-arcsq] (V11) -- (V3) node[pos=0.5,sloped,above] {};
      \draw (0, 1.85) circle (0.05);
      \draw[double arrow] (V8) -- (V5)
       node[pos=0.5,sloped,above] {};
      \draw[-arcsq] (V7) -- (V8) node[pos=0.5,sloped,above] {};
      \draw (3.0, 0.85) circle (0.05);
      \draw[-arcsq] (V12) -- (V8) node[pos=0.5,sloped,above] {};
      \draw (3.95, 0.0) circle (0.05);
      \draw (2.1, 1.8) circle (0.05);
      \draw[-, color=red] (V2) -- (V7) node[pos=0.5,sloped,above] {};
      \draw [color=red] (2.85, 1.2) circle (0.05);
      \draw (2.0, -0.7) node(con3) [] {{\small\,(d) $\mathcal{L}_{\mathit{MMB^{+}}(V_4)}$\,}};
        \end{tikzpicture}
    \end{subfigure}
     \hspace{0.5em}
    \begin{subfigure}{0.3\textwidth}
    \begin{tikzpicture}[node distance=1.5cm and 2cm, thick]
      \draw (0.0, 2.0) node(V11) [] {{\scriptsize\,$V_{11}$\,}};
      \draw (0.0, 0.0) node(V10) [] {{\scriptsize\,$V_{10}$\,}};
      \draw (1.5, 0.0) node(V5) [] {{\scriptsize\,$V_{5}$\,}};
      \draw (3.0, 0.0) node(V8) [] {{\scriptsize\,$V_{8}$\,}};
      \draw (4.3, 0.0) node(V12) [] {{\scriptsize\,$V_{12}$\,}};
      \draw (0.0, 1.0) node(V3) [] {{\scriptsize\,$V_{3}$\,}};
      \draw (1.5, 1.0) node(V4) [] {{\scriptsize\,$V_{4}$\,}};
      \draw (3.0, 1.0) node(V7) [] {{\scriptsize\,$V_{7}$\,}};
      \draw (2.25, 2.0) node(V2) [] {{\scriptsize\,$V_{2}$\,}};
      
      \draw[double arrow] (V3) -- (V4);
      \draw[-arcsq] (V2) -- (V4) node[pos=0.5,sloped,above] {};
      
      \draw[-arcsq] (V4) -- (V5) node[pos=0.5,sloped,above] {};
      \draw[-arcsq] (V11) -- (V3) node[pos=0.5,sloped,above] {};
      \draw (0, 1.85) circle (0.05);
      \draw[-arcsq] (V5)  --  (V10) node[pos=0.5,sloped,above] {};
      \draw[double arrow] (V8) -- (V5)
       node[pos=0.5,sloped,above] {};
      \draw[-arcsq] (V7) -- (V8) node[pos=0.5,sloped,above] {};
      \draw (3.0, 0.85) circle (0.05);
      \draw[-arcsq] (V12) -- (V8) node[pos=0.5,sloped,above] {};
      \draw (3.95, 0.0) circle (0.05);
      \draw (2.1, 1.8) circle (0.05);
      \draw (0.0, 0.8) circle (0.05);
      \draw[-arcsq] (V3) -- (V10) node[pos=0.5,sloped,above] {};
      \draw (2.0, -0.7) node(con3) [] {{\small\,(e) Updated $\mathcal{P}$ after learning $\mathcal{L}_{\mathit{MMB^{+}}(V_4)}$\,}};
        \end{tikzpicture}
    \end{subfigure}
     \hspace{0.5em}
    \begin{subfigure}{0.3\textwidth}
      \begin{tikzpicture}[node distance=1.5cm and 2cm, thick]
      \draw (0.0, 2.0) node(V11) [] {{\scriptsize\,$V_{11}$\,}};
      \draw (0.0, 0.0) node(V10) [] {{\scriptsize\,$V_{10}$\,}};
      \draw (1.5, 0.0) node(V5) [] {{\scriptsize\,$V_{5}$\,}};
      \draw (3.0, 0.0) node(V8) [] {{\scriptsize\,$V_{8}$\,}};
      \draw (4.3, 0.0) node(V12) [] {{\scriptsize\,$V_{12}$\,}};
      \draw (0.0, 1.0) node(V3) [] {{\scriptsize\,$V_{3}$\,}};
      \draw (1.5, 1.0) node(V4) [] {{\scriptsize\,$V_{4}$\,}};
      \draw (3.0, 1.0) node(V7) [] {{\scriptsize\,$V_{7}$\,}};
      \draw (2.25, 2.0) node(V2) [] {{\scriptsize\,$V_{2}$\,}};
      
      \draw[double arrow] (V3) -- (V4) node[pos=0.5,sloped,above] {};
      
      \draw[-arcsq] (V2) -- (V4) node[pos=0.5,sloped,above] {};
      \draw[-arcsq, color=blue] (V4) -- (V5) node[pos=0.5,sloped,above] {};
      \draw[-arcsq] (V11) -- (V3) node[pos=0.5,sloped,above] {};
      \draw (0, 1.85) circle (0.05);
      \draw[-arcsq, color=blue] (V5)  --  (V10) node[pos=0.5,sloped,above] {};
      \draw[double arrow, color=blue] (V8) -- (V5)
       node[pos=0.5,sloped,above] {};
      \draw[-arcsq] (V7) -- (V8) node[pos=0.5,sloped,above] {};
      \draw (3.0, 0.85) circle (0.05);
      \draw[-arcsq] (V12) -- (V8) node[pos=0.5,sloped,above] {};
      \draw (3.95, 0.0) circle (0.05);
      \draw (2.1, 1.8) circle (0.05);
      \draw (0.0, 0.8) circle (0.05);
      \draw[-arcsq] (V3) -- (V10) node[pos=0.5,sloped,above] {};
      
      \draw (2.0, -0.7) node(con3) [] {{\small\,(f) The Final local $\mathcal{P}$\,}};
        \end{tikzpicture}
    \end{subfigure}
    \vspace{-3mm}
    \caption{The sequential process for finding the parents and children of the target $V_{5}$ in the graph of \autoref{fig-main-example} (a), where the red edges indicate that the current local results cannot be guaranteed to be consistent with the global learning results.}
    \label{fig:inferred local PAG process}
    \vspace{-2mm}
\end{figure*}
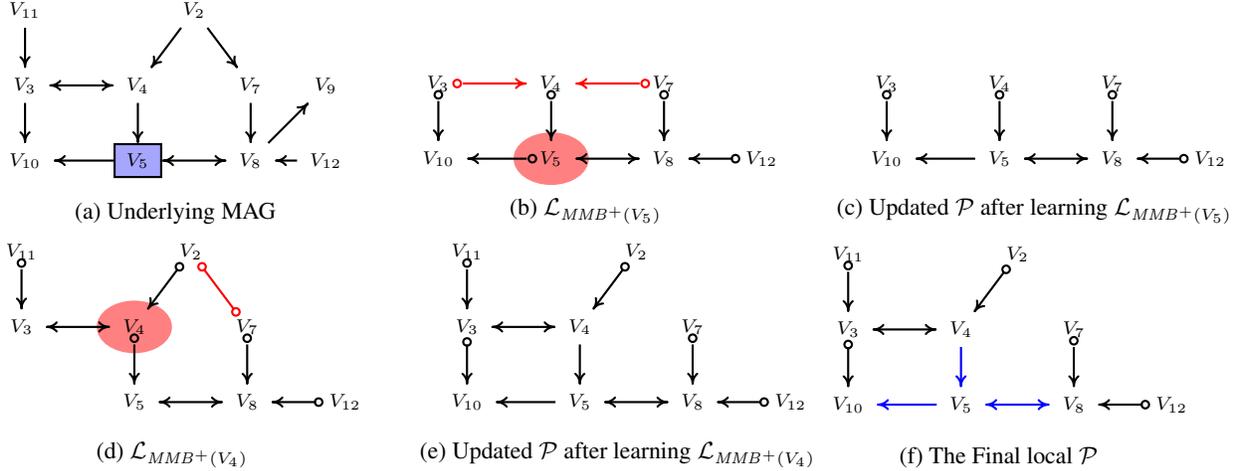
In this section, we leverage the above theoretical results and propose a sequential algorithm to learn the potential parents and children of a target node in models that include latent variables.
{We use $\mathcal{L}_{\mathcal{V}}$ to denote the local structure learned from a subset $\mathcal{V}$ of $\mathbf{V}$, utilizing the test of conditional independence and orientation of V-structures}.
We use $\mathcal{P}$ to store the local causal structure around $T$, where the causal information preserved in this structure is consistent with the global learning structure.
Roughly speaking, for target $T$, our method contains the following key steps:
\begin{description}[leftmargin=14pt,itemsep=0pt,topsep=0pt,parsep=0pt]
    \item [S1.] Finding a MAG Markov blanket $\mathit{MMB(T)}$ of the target $T$ and learning the local structure $\mathcal{L}_{\mathit{MMB}^{+}(T)}$. 
    \item [S2.] Putting the edges connected to $T$ and the V-structures containing $T$ in $\mathcal{L}_{\mathit{MMB}^{+}(T)}$ to $\mathcal{P}$, according to \autoref{theorem1} and \autoref{theorem2}.
    \item [S3.] Orienting maximally the edge marks in $\mathcal{P}$ using the standard orientation rules of \citet{zhang2008completeness}.
\end{description}
Three steps S1 $\sim$ S3 are repeated sequentially until any one of the stop rules is met in \autoref{theorem3}. For notational convenience, let $\mathbf{WaitList}$ be the list of nodes to be checked by \autoref{theorem1} and \autoref{theorem2}. Let $\mathbf{DoneList}$ denote the list of nodes whose local structures have been found.
Additionally, $\mathit{MMB_{alg}}$ refers to the algorithm used for learning $\mathit{MMB}$. The complete procedure is summarized in Algorithm \autoref{MMB-by-MMB} and \autoref{MMB-alg}, and a complete example is given in Example \ref{example-set-completed-algorithm}. 


\begin{example-set}\label{example-set-completed-algorithm}
We illustrate our MMB-by-MMB algorithm with the causal MAG in \autoref{fig:inferred local PAG process}(a). We assume Oracle tests for conditional independence tests. In this structure, There are latent variables between nodes $V_3$ and $V_4$, $V_5$ and $V_8$. We here are interested in the local structure around $V_5$.
The learning process is as follows:
\begin{itemize}[leftmargin=14pt,itemsep=0pt,topsep=0pt,parsep=0pt]
    \item It first initializes sets $\mathbf{Waitlist}$ = $\{ V_5\}$, $\mathbf{Donelist}$ = $\emptyset$, and graph $\mathcal{P}=\emptyset$ (Line 1).
    \item After initialization, it runs $\mathit{MMB_{alg}(V_5)}$ and obtain $\mathit{MMB(V_5)}\!\!=\!\!\{V_3, V_4, V_7, V_8, V_{10}, V_{12}\}$ (Lines $3 \sim 4$). 
    \item It then learns $\mathcal{L}_{\mathit{MMB^{+}(V_5)}}$ over $\mathit{MMB^{+}(V_5)}$, as depicted in \autoref{fig:inferred local PAG process}(b) (Line 10).
    \item Next, it updates $\mathcal{P}$ by selecting the edges connected to $V_5$ and the V-structures containing $V_5$ (Line 12).
    \item It now orients $V_5 \circ\!\!\rightarrow V_{10}$ as $V_5 \rightarrow V_{10}$ by orientation rules (Line 13). Consequently, it obtains $\mathcal{P}$ as shown in \autoref{fig:inferred local PAG process}(c).
    \item Then, it updates $\mathbf{Donelist}\!\!=\!\!\{V_5\}$, and $\mathbf{Waitlist}\!\!=\!\!\{V_4, V_8, V_{10}\}$ (Lines $14 \sim 15$).
    \item Sequentially, it runs $\mathit{MMB_{alg}(V_4)}$ and obtain $\mathit{MMB}(V_4)\!\!=\!\!\{V_2,V_3,V_5,V_7,V_8,V_{11},V_{12}\}$.
    \item It then learns the local structure $\mathcal{L}_{\mathit{MMB^{+}(V_4)}}$ as shown in \autoref{fig:inferred local PAG process}(d) (Line 10).
    \item Next, it pools the determined edges together and orients $V_4 \circ\!\!\rightarrow V_5$ as $V_4 \rightarrow V_5$ (Lines $12 \sim 13$). Following this, it derives the local structure $\mathcal{P}$, as illustrated in \autoref{fig:inferred local PAG process}(e).
    \item Next, it updates sets $\mathbf{Waitlist}\!\!=\!\!\{V_8,V_{10},V_2,V_3\}$, $\mathbf{Donelist}\!\!=\!\!\{V_5, V_4\}$ (Lines $14 \sim 15$).
    \item Finally, the algorithm terminates because stop $\mathcal{R}1$ is satisfied. Output the resulting local structure $\mathcal{P}$, which is depicted in \autoref{fig:inferred local PAG process}(f).
\end{itemize}
More details of the example are given in the Appendix \ref{Appendix-Algorithm}.
\end{example-set}

\begin{theorem}\label{theorem-4}
    \textbf{The Correctness of MMB-by-MMB Algorithm.} 
    We assume Oracle tests for conditional independence tests and accurately obtain the MMB of the target variable $T$ through the MMB discovery algorithm, 
    MMB-by-MMB will identify the direct causes and effects of the target under causal faithfulness and no selection bias assumptions.
\end{theorem}

\autoref{theorem-4} shows that the local $\mathcal{P}$ obtained through the MMB-by-MMB algorithm is correct, meaning that the edges connected to the target node $T$ and their directions in the output $\mathcal{P}$ are identical to those in the Markov equivalence class of the underlying causal MAG $\mathcal{M}$.
    
\subsection{Complexity of MMB-by-MMB Algorithm}\label{Subsec-Complexity}

This algorithm's complexity can be divided into two parts: the first part involves finding the MAG Markov blanket, and the second part involves learning the local structure. Let $r$ denote the number of local structures to be learned sequentially in our algorithm. In our experiment, we used the TC algorithm\cite{pellet2008using} to search for MMB. The time complexity of finding MMB among $r$ nodes out of $n$ total nodes is $\mathcal{O} \left( \frac{r(2n-r-1)}{2} \right)$, where $\mathit{n}$ denotes the size of observed node set $\mathbf{O}$. When learning local structure, we apply the logic of the PC algorithm to identify the adjacent edges in $\mathit{MMB^{+}(T)}$. In the worst case, the complexity of the PC algorithm for learning a local structure over $n$ nodes is $\mathcal{O} \left( 2\binom{n}{2} \sum_{i=0}^{k}\binom{n-1}{i}  \right)$. Let $\mathit{k}$ be the maximal degree of any variable and let $\mathit{|MMB^{+}|}$ denote the size of $\mathit{MMB^{+}(V_X)}$.  The complexity of the MMB-by-MMB algorithm is the total complexity of finding MAG Markov blankets plus constructing local structures. In the worst case, this is, $\mathcal{O}\left[\frac{r(2n-r-1)}{2} + 2r\binom{|MMB^{+}|}{2} \sum_{i=0}^{k}\binom{|MMB^{+}|-1}{i}   \right]$.  

\begin{table*}[hpt!]\scriptsize
\centering
\small
\caption{Performance Comparisons on MILDEW.Net}
\resizebox{0.98\textwidth}{!}{
\begin{tabular}{cc|ccccc|ccccc}
\toprule
\multicolumn{2}{c}{} & \multicolumn{5}{|c|}{Size=1000} & \multicolumn{5}{c}{Size=5000}\\
\hline
\multicolumn{1}{c}{Target} & Algorithm & {\textbf{Precision}$\uparrow$} & {\textbf{Recall}$\uparrow$} &{\textbf{F1}$\uparrow$} & {\textbf{Distance}$\downarrow$} & {\textbf{nTest}$\downarrow$} & {\textbf{Precision}}$\uparrow$ & {\textbf{Recall}$\uparrow$} &{\textbf{F1}$\uparrow$} & {\textbf{Distance}$\downarrow$} & {\textbf{nTest}$\downarrow$}\\
\hline
& PC-stable    
& 0.28±0.08 & 0.28±0.09 & 0.28±0.08 & 1.02±0.12 & 5032.70 
& 0.27±0.07 & 0.27±0.07 & 0.27±0.07 & 1.04±0.10 & 8930.25   \\
& FCI        
& 0.83±0.27 & 0.83±0.27 & 0.83±0.27 & 0.24±0.39 & 10260.86
& 0.85±0.27 & 0.85±0.26 & 0.85±0.27 & 0.21±0.38 & 16637.09  \\
& RFCI       
& 0.81±0.28 & 0.81±0.28 & 0.81±0.28 & 0.27±0.39 & 5032.70  
& 0.84±0.26 & 0.84±0.27 & 0.84±0.27 & 0.22±0.38 & 8930.25   \\
{dm-1}& MB-by-MB   
& 0.51±0.13 & 0.57±0.18 & 0.52±0.14 & 0.67±0.20 & 4596.90 
& 0.50±0.12 & 0.58±0.18 & 0.51±0.13 & 0.68±0.19 & 15864.89 \\
& CMB        
& 0.49±0.15 & 0.52±0.17 & 0.50±0.15 & 0.71±0.22 & 3440.64  
& 0.50±0.13 & 0.54±0.16 & 0.51±0.14 & 0.69±0.20 & 3661.29  \\
& GraN-LCS  
& 0.70±0.22 & 0.78±0.20 & 0.73±0.22 & 0.39±0.30 & -
& 0.67±0.21 & 0.74±0.20 & 0.69±0.20 & 0.43±0.29 & -                \\
& MMB-by-MMB 
& \textbf{0.95±0.15} & \textbf{0.95±0.15} & \textbf{0.95±0.15} & \textbf{0.07±0.22} & \textbf{392.95}    
& \textbf{0.97±0.13} & \textbf{0.97±0.13} & \textbf{0.97±0.13} & \textbf{0.04±0.18} & \textbf{613.61}   \\
\hline
& PC-stable   
& 0.33±0.18   & 0.33±0.18   & 0.33±0.18   & 0.94±0.26   & 5071.66    
& 0.33±0.18   & 0.33±0.18   & 0.33±0.18   & 0.95±0.25   & 8972.31     \\
& FCI        
& 0.77±0.20   & 0.70±0.20   & 0.72±0.20   & 0.39±0.29   & 10338.28 
& 0.84±0.15   & 0.77±0.20   & 0.80±0.18   & 0.29±0.26   & 16762.44    \\
& RFCI       
& 0.72±0.21   & 0.64±0.19   & 0.67±0.20   & 0.47±0.28   & 5071.66    
& 0.81±0.15   & 0.73±0.20   & 0.76±0.18   & 0.35±0.26   & 8972.31    \\
{dm-4}& MB-by-MB   
& 0.59±0.14 & 0.69±0.18 & 0.62±0.14 & 0.53±0.20 & 7075.49
& 0.60±0.15 & 0.71±0.18 & 0.63±0.16 & 0.51±0.22 & 28815.65 \\
& CMB        
& 0.60±0.15   & 0.63±0.16   & 0.61±0.15   & 0.55±0.22   & 2325.96  
& 0.58±0.15   & 0.63±0.15   & 0.60±0.15   & 0.57±0.21   & 3638.17    \\
& GraN-LCS   
& 0.57±0.17 & 0.59±0.19 & 0.56±0.16 & 0.62±0.23 & - 
& 0.60±0.20 & 0.61±0.21 & 0.59±0.19 & 0.58±0.26 & -  \\
& MMB-by-MMB 
& \textbf{0.95±0.13}   & \textbf{0.91±0.16}   & \textbf{0.92±0.15}   & \textbf{0.11±0.21}   & \textbf{527.14}    
& \textbf{0.99±0.05}   & \textbf{0.98±0.09}   & \textbf{0.98±0.08}   & \textbf{0.03±0.11}   & \textbf{690.57}  \\
\bottomrule
\end{tabular}
}
\label{table-1}
\begin{tablenotes}
		\item \scriptsize{Note: The symbol '-' indicates that GraN-LCS does not output this information. MMB-by-MMB with the second best result is underlined. $\uparrow$ means a higher value is better, and vice versa.}
\end{tablenotes}
\vspace{-4mm}
\end{table*}

\begin{table*}[hpt!]
\centering
\small
\caption{Performance Comparisons on ALARM.Net}
\resizebox{0.98\textwidth}{!}{
\begin{tabular}{cc|ccccc|ccccc}
\toprule
\multicolumn{2}{c}{} & \multicolumn{5}{|c|}{Size=1000} & \multicolumn{5}{c}{Size=5000}\\
\hline
\multicolumn{1}{c}{Target} & Algorithm & {\textbf{Precision}$\uparrow$} & {\textbf{Recall}$\uparrow$} &{\textbf{F1}$\uparrow$} & {\textbf{Distance}$\downarrow$} & {\textbf{nTest}$\downarrow$} & {\textbf{Precision}}$\uparrow$ & {\textbf{Recall}$\uparrow$} &{\textbf{F1}$\uparrow$} & {\textbf{Distance}$\downarrow$} & {\textbf{nTest}$\downarrow$}\\
\hline
& PC-stable 
& 0.56±0.16 & 0.55±0.14 & 0.56±0.15 & 0.63±0.20 &3515.85  
& 0.56±0.16 & 0.55±0.14 & 0.56±0.15 & 0.63±0.20 &4878.73  \\
& FCI 
& 0.78±0.25 & 0.77±0.25 & 0.77±0.25 & 0.32±0.35 & 7552.66 
& 0.84±0.24 & 0.83±0.24 & 0.83±0.24 & 0.24±0.34 & 10173.30 \\
& RFCI 
& 0.78±0.25 & 0.77±0.24 & 0.77±0.24 & 0.32±0.34 & 3515.85  
& 0.83±0.23 & 0.83±0.24 & 0.83±0.24 & 0.24±0.33 & 4878.73   \\
{LVEDVOLUME} & MB-by-MB 
& 0.46±0.18 & 0.46±0.18 & 0.45±0.17 & 0.77±0.25 & 1531.45
& 0.44±0.15 & 0.44±0.16 & 0.43±0.14 & 0.81±0.20 & 4196.87 \\
& CMB 
& 0.44±0.21 & 0.43±0.19 & 0.43±0.20 & 0.80±0.28 & 1471.71 
& 0.44±0.20 & 0.43±0.18 & 0.43±0.18 & 0.81±0.26 & 1992.32 \\
& GraN-LCS 
& 0.58±0.15 & 0.57±0.14 & 0.57±0.14 & 0.61±0.20 & -     
& 0.58±0.13 & 0.58±0.13 & 0.58±0.13 & 0.60±0.18 & -  \\
&MMB-by-MMB 
& \textbf{0.97±0.12} & \textbf{0.96±0.12} & \textbf{0.96±0.12} & \textbf{0.05±0.16} & \textbf{324.09} 
& \textbf{0.98±0.09} & \textbf{0.98±0.09} & \textbf{0.98±0.09} & \textbf{0.03±0.12} & \textbf{344.51} \\
\hline
& PC-stable 
& 0.57±0.17 & 0.56±0.13 & 0.56±0.14 & 0.62±0.20 & 3470.71
& 0.57±0.17 & 0.56±0.13 & 0.56±0.14 & 0.62±0.20 & 4840.23\\
& FCI 
& 0.78±0.25 & 0.77±0.25 & 0.77±0.25 & 0.33±0.35 & 7483.00 
& 0.84±0.23 & 0.84±0.23 & 0.84±0.23 & 0.23±0.33 & 10171.28  \\
& RFCI 
& 0.72±0.25 & 0.71±0.24 & 0.71±0.25 & 0.41±0.35 & 3470.71  
& 0.83±0.24 & 0.83±0.24 & 0.83±0.24 & 0.24±0.34 & 4840.23 \\
{STROKEVOLUME} & MB-by-MB 
& 0.43±0.18 & 0.48±0.21 & 0.43±0.18 & 0.79±0.25 & 2076.76
& 0.37±0.15 & 0.41±0.17 & 0.38±0.15 & 0.88±0.21 & 6377.99\\
& CMB 
& 0.44±0.20 & 0.45±0.20 & 0.44±0.19 & 0.79±0.27 & 2197.08
& 0.37±0.19 & 0.37±0.19 & 0.37±0.19 & 0.89±0.26 & 2597.72\\
& GraN-LCS 
& 0.42±0.13 & 0.46±0.16 & 0.43±0.14 & 0.80±0.20 & -     
& 0.42±0.13 & 0.46±0.16 & 0.43±0.14 & 0.80±0.20 & -     \\
&MMB-by-MMB 
& \textbf{0.95±0.15} & \textbf{0.95±0.16} & \textbf{0.95±0.16} & \textbf{0.08±0.22} & \textbf{566.39} 
& \textbf{0.98±0.09} & \textbf{0.98±0.09} & \textbf{0.98±0.09} & \textbf{0.02±0.12} & \textbf{698.17} \\
\bottomrule
\end{tabular}
}
\label{table-2}
\begin{tablenotes}
\item \scriptsize{Note: The symbol '-' indicates that GraN-LCS does not output this information. MMB-by-MMB with the second best result is underlined. $\uparrow$ means a higher value is better, and vice versa.}
\end{tablenotes}
\vspace{-4mm}
\end{table*}

\begin{table*}[hpt!]
\centering
\small
\caption{Performance Comparisons on WIN95PTS.Net}
\resizebox{0.98\textwidth}{!}{
\begin{tabular}{cc|ccccc|ccccc}
\toprule
\multicolumn{2}{c}{} & \multicolumn{5}{|c|}{Size=1000} & \multicolumn{5}{c}{Size=5000}\\
\hline
\multicolumn{1}{c}{Target} & Algorithm & {\textbf{Precision}$\uparrow$} & {\textbf{Recall}$\uparrow$} &{\textbf{F1}$\uparrow$} & {\textbf{Distance}$\downarrow$} & {\textbf{nTest}$\downarrow$} & {\textbf{Precision}}$\uparrow$ & {\textbf{Recall}$\uparrow$} &{\textbf{F1}$\uparrow$} & {\textbf{Distance}$\downarrow$} & {\textbf{nTest}$\downarrow$}\\
\hline
&Pc-stable           
& 0.52±0.07 & 0.52±0.07 & 0.52±0.07 & 0.68±0.10 & 12657.62    
& 0.53±0.07 & 0.53±0.08 & 0.53±0.07 & 0.67±0.10 & 26398.40 \\
&FCI                
& 0.69±0.24 & 0.69±0.24 & 0.68±0.24 & 0.45±0.34 & 25417.23  
& 0.77±0.27 & 0.76±0.27 & 0.76±0.27 & 0.34±0.38 & 43850.55 \\
&RFCI               
& 0.67±0.23 & 0.66±0.23 & 0.66±0.23 & 0.48±0.33 & 12657.62    
& 0.77±0.27 & 0.75±0.28 & 0.76±0.28 & 0.35±0.39 & 26398.40 \\
{Problem5}&MB-by-MB           
& 0.46±0.19 & 0.50±0.22 & 0.47±0.20 & 0.75±0.28 & 13633.52 
&NA &NA &NA &NA &NA \\
&CMB                
& 0.57±0.16 & 0.60±0.18 & 0.58±0.17 & 0.59±0.23  & 4757.95    
& 0.54±0.15 & 0.58±0.17 & 0.56±0.16 & 0.63±0.22 & \textbf{5413.78}   \\
&GraN-LCS           
& 0.48±0.14 & 0.50±0.15 & 0.48±0.14 & 0.73±0.20 & -         
& 0.48±0.16 & 0.48±0.17 & 0.48±0.16 & 0.74±0.22 & -      \\
&MMB-by-MMB 
& \textbf{0.90±0.20} & \textbf{0.90±0.20} & \textbf{0.89±0.20} & \textbf{0.15±0.29} & \textbf{3907.42}     
& \textbf{0.93±0.18} & \textbf{0.92±0.19} & \textbf{0.92±0.19} & \textbf{0.11±0.27} & \underline{12372.20} \\
\hline
& PC-stable 
& 0.78±0.08 & 0.77±0.06 & 0.78±0.07 & 0.32±0.10 & 12637.44 
& 0.77±0.06 & 0.76±0.04 & 0.76±0.05 & 0.34±0.07 & 25058.00  \\
& FCI 
& 0.80±0.24 & 0.80±0.24 & 0.80±0.24 & 0.28±0.34 & 25651.52 
& 0.83±0.24 & 0.83±0.25 & 0.83±0.25 & 0.24±0.35 & 42604.71  \\
& RFCI 
& 0.81±0.24 & 0.81±0.24 & 0.81±0.24 & 0.27±0.34 & 12637.44 
& 0.82±0.25 & 0.82±0.25 & 0.82±0.25 & 0.25±0.35 & 25058.00  \\
{HrglssDrtnAftrPrnt} & MB-by-MB 
& 0.46±0.09 & 0.46±0.09 & 0.46±0.09 & 0.77±0.13 & 14169.84 
& 0.45±0.11 & 0.45±0.11 & 0.45±0.11 & 0.78±0.15 & 45118.80 \\
& CMB 
& 0.48±0.23 & 0.48±0.23 & 0.48±0.23 & 0.73±0.33 & 7933.74  
& 0.42±0.19 & 0.42±0.19 & 0.42±0.19 & 0.82±0.26 & 11783.44 \\
& GraN-LCS 
& 0.39±0.14   & 0.39±0.14   & 0.39±0.14   & 0.86±0.20   & -         
& 0.43±0.13   & 0.43±0.13   & 0.43±0.13   & 0.80±0.18   & -   \\
&MMB-by-MMB 
& \textbf{0.92±0.18} & \textbf{0.92±0.18} & \textbf{0.92±0.18} & \textbf{0.11±0.25} & \textbf{1054.52} 
& \textbf{0.92±0.20} & \textbf{0.92±0.20} & \textbf{0.92±0.20} & \textbf{0.11±0.29} & \textbf{2029.79}   \\
\bottomrule
\end{tabular}
}
\label{table-3}
\begin{tablenotes}
		\item \scriptsize{Note: The symbol '-' indicates that GraN-LCS does not output this information. MMB-by-MMB with the second best result is underlined. $\uparrow$ means a higher value is better, and vice versa. NA entries for MB-by-MB demonstrate that the runtime exceeds a certain threshold.}
\end{tablenotes}
\vspace{-4mm}
\end{table*}


\begin{table*}[hpt!]
\centering
\small
\caption{Performance Comparisons on ANDES.Net}
\resizebox{0.98\textwidth}{!}{
\begin{tabular}{cc|ccccc|ccccc}
\toprule
\multicolumn{2}{c}{} & \multicolumn{5}{|c|}{Size=1000} & \multicolumn{5}{c}{Size=5000}\\
\hline
\multicolumn{1}{c}{Target} & Algorithm & {\textbf{Precision}$\uparrow$} & {\textbf{Recall}$\uparrow$} &{\textbf{F1}$\uparrow$} & {\textbf{Distance}$\downarrow$} & {\textbf{nTest}$\downarrow$} & {\textbf{Precision}}$\uparrow$ & {\textbf{Recall}$\uparrow$} &{\textbf{F1}$\uparrow$} & {\textbf{Distance}$\downarrow$} & {\textbf{nTest}$\downarrow$}\\
\hline
& PC-stable 
&0.23±0.07 & 0.23±0.07 & 0.23±0.07 & 1.09±0.10 & 234677.37 
& 0.23±0.07 & 0.23±0.07 & 0.23±0.07 & 1.09±0.10 & 439483.08  \\
& FCI 
& 0.70±0.24 & 0.68±0.25 & 0.68±0.24 & 0.45±0.34 & 901063.34 
& 0.79±0.24 & 0.78±0.25 & 0.79±0.24 & 0.30±0.34 & 1584682.77  \\
& RFCI 
& 0.66±0.24 & 0.64±0.24 & 0.65±0.24 & 0.50±0.34 & 234677.37  
& 0.78±0.24 & 0.77±0.25 & 0.77±0.25 & 0.32±0.35 & 439483.08   \\
{RApp3($V_5$)} & MB-by-MB 
& 0.34±0.07 & 0.43±0.12 & 0.36±0.08 & 0.89±0.12 & 24239.83 
& 0.39±0.08 & 0.56±0.12 & 0.43±0.09 & 0.78±0.12 & 44225.00\\
& CMB 
& 0.33±0.06 & 0.38±0.09 & 0.34±0.07 & 0.92±0.09 & 79932.47 
& 0.32±0.05 & 0.39±0.09 & 0.34±0.07 & 0.93±0.09 & 145631.64 \\
& GraN-LCS 
& 0.39±0.12 & 0.46±0.16 & 0.40±0.12 & 0.84±0.17 & -           
& 0.38±0.13 & 0.42±0.16 & 0.39±0.13 & 0.86±0.19 & -              \\
&MMB-by-MMB 
& \textbf{0.91±0.15} & \textbf{0.90±0.17} & \textbf{0.89±0.17} & \textbf{0.16±0.24} & \textbf{5043.44}
& \textbf{0.98±0.07} & \textbf{0.98±0.07} & \textbf{0.98±0.07} & \textbf{0.03±0.10} & \textbf{4595.15}  \\

\hline
& PC-stable 
& 0.46±0.09 & 0.46±0.09 & 0.46±0.09 & 0.77±0.13 & 234677.37 
& 0.46±0.09 & 0.46±0.09 & 0.46±0.09 & 0.77±0.13 & 439483.08  \\
& FCI 
& 0.79±0.24 & 0.78±0.24 & 0.78±0.24 & 0.32±0.34 & 901063.34 
& 0.84±0.23 & 0.84±0.23 & 0.84±0.24 & 0.23±0.33 & 1584682.77  \\
& RFCI 
& 0.79±0.24 & 0.79±0.24 & 0.79±0.24 & 0.32±0.34 & 234677.37 
& 0.84±0.23 & 0.83±0.23 & 0.83±0.24 & 0.24±0.33 & 439483.08  \\
{RApp4} & MB-by-MB 
& 0.26±0.14 & 0.29±0.16 & 0.26±0.14 & 1.04±0.20 & 18761.00
&NA &NA &NA &NA &NA \\
& CMB 
& 0.25±0.16 & 0.26±0.17 & 0.25±0.16 & 1.06±0.23 & 87956.70
& 0.23±0.12 & 0.23±0.12 & 0.23±0.12 & 1.09±0.17 & 212387.71 \\
& GraN-LCS 
& 0.36±0.16 & 0.45±0.19 & 0.38±0.16 & 0.87±0.23 & -        
& 0.38±0.17 & 0.48±0.22 & 0.40±0.18 & 0.84±0.26 & -       \\
&MMB-by-MMB 
& \textbf{0.91±0.22} & \textbf{0.93±0.19} & \textbf{0.91±0.21} & \textbf{0.12±0.30} & \textbf{3153.11}     
& \textbf{0.97±0.12} & \textbf{0.98±0.09} & \textbf{0.97±0.11} & \textbf{0.04±0.15} & \textbf{1430.12}  \\

\bottomrule
\end{tabular}
}
\label{table-4}
\begin{tablenotes}
		\item \scriptsize{Note: The symbol '-' indicates that GraN-LCS does not output this information. MMB-by-MMB with the second best result is underlined. $\uparrow$ means a higher value is better, and vice versa. NA entries for MB-by-MB demonstrate that the runtime exceeds a certain threshold.}
\end{tablenotes}
\end{table*}
\section{Experimental Result}
To demonstrate the accuracy and efficiency of our algorithm, we compared the proposed MMB-by-MMB algorithm with global learning methods, such as PC-stable~\cite{colombo2014order}, FCI~\cite{spirtes2000causation}, and RFCI~\cite{colombo2012learning} \footnote{For PC-stable algorithms, we used the implementations in the MATLAB package
at \url{https://github.com/kuiy/CausalLearner}. FCI algorithm is from Python-package causallearn \cite{zheng2023causal}. RFCI algorithm is from R-package pcalg \cite{kalisch2012causal}.}. We also compared with local learning methods, such as the MB-by-MB algorithm~\cite{wang2014discovering}, the Causal Markov Blanket (CMB) algorithm~\cite{gao2015local}, and the GradieNt-based LCS (GraN-LCS) algorithm~\cite{liang2023gradient} 
\footnote{We utilized the Python pyCausalFS package \cite{yu2020causality} for MB-by-MB and CMB algorithms. The source code is available at \url{https://github.com/kuiy/pyCausalFS}, and the GraN-LCS algorithm 
from~\url{https://www.sdu-idea.cn/codes.php?name=GraN-LCS}.}.

We here use the existing implementation \cite{pellet2008using} of the Total Conditioning (TC) discovery algorithm to find the MB of a target variable.
Our source code is available from \url{https://github.com/fengxie009/MMB-by-MMB}.

\subsection{Synthetic Data Generated from Benchmark Network Structures}\label{Subsection-Synthetic-Data}
\textbf{\emph{Experimental setup:}}
We select four networks ranging from low to high dimensionality:  MILDEW, ALARM,  WIN95PTS, and ANDES, containing 35, 37, 76, and 223 nodes, respectively\footnote{The details of those networks can be found at \url{https://www.bnlearn.com/bnrepository/}.}.
The network structures are parameterized as a linear Gaussian structural causal model.
The causal strength of each edge is drawn from Uniform $([-1,-0.5]\cup[0.5,1])$. For each graph, we randomly select {4, 4, 6,10} nodes as latent variables, and others as observed variables. We here choose nodes with more adjacent nodes as target nodes. Each experiment was repeated 100 times with randomly generated data, and the reported results were averaged. The best results are highlighted in boldface.

\emph{\textbf{Metrics:}} 
We evaluate the performance of the algorithms using
the following typical metrics: \textbf{Precision}: the ratio of true edges \footnote{A true edge implies the correct estimation of tails on both sides.} in the output to the total number of edges in the algorithm's output. 
\textbf{Recall}: the ratio of true edges in the output to the total number of edges in the ground-truth structure of a target.
\textbf{F1}: the harmonic average of \textit{Precision} and \textit{Recall}, calculated as
\begin{nospaceflalign}\nonumber
    {\mathit{F1} = 2*\mathit{Precision}*\mathit{Recall}/(\mathit{Prescision} + \mathit{Recall}).}
\end{nospaceflalign}
\textbf{Distance}: the Euclidean distance between Recall and Precision, computed as
\begin{nospaceflalign}\nonumber
    {\mathit{Distance} = \sqrt{(1 - \mathit{Recall})^2 + (1 - \mathit{Precision})^2}.}
\end{nospaceflalign}
\textbf{nTest}: the number of conditional independence tests implemented by an algorithm.

\emph{\textbf{Results:}} Due to space constraints, we here present only partial results for each network with two targets. These results are shown in Tables \ref{table-1} $\sim$ \ref{table-4}. The complete results are given in the Appendix \ref{Appendix-simulations}. From the tables, we can see that our proposed MMB-by-MMB algorithm outperforms other methods with almost all evaluation metrics in all four structures and in all sample sizes, indicating the effectiveness of our method. As expected, the number of conditional independence tests in our method is far less than that in the methods FCI and RFCI, which are used for global learning structures involving latent variables.
It is worth noting that although the \emph{nTest} of CMB method is fewer than our method in the WIN95PTS network when $Size=5000$, the other four metrics of our method outperform CMB. Furthermore, the results of local learning methods, i.e., the MB-by-MB, CMB, and GraN-LCS, are not satisfactory, indicating that these methods cannot address situations involving latent variable structures.

\subsection{Gene Expression Data}
In this section, we apply our method on the gene expression data from \citet{wille2004sparse}, which comprises gene expression measurements of Arabidopsis thaliana grown under 118 different conditions, such as variations in light and darkness, and exposure to growth hormones. \citet{wille2004sparse} focused particularly on the genes involved in isoprenoid synthesis. In Arabidopsis thaliana, isoprenoid synthesis is carried out by two distinct pathways in separate organs: the mevalonate pathway (MVA) and the nonmevalonate pathway (MEP). The dataset we used contains 33 genes. We here employ the model in \citet{wille2004sparse} as a baseline (See Figure 3 of~\citet{wille2004sparse}). It should be noted that in this model, some edges are undirected.
We selected two genes, \emph{DXR} and \emph{MECPS}, as target nodes, respectively. The findings are as follows:

\textbf{Target=DXR.} 
Our method obtains that $\mathit{Pa}(\mathit{DXR})=\{\mathit{HMGS}\}$ and $\mathit{Ch}(\mathit{DXR})$=\emph{\{DXPS2, CMK, MECPS, HDS\}}.
We found that the connections among four genes \emph{DXPS2}, \emph{CMK}, \emph{MECPS}, and \emph{HDS}, as well as the information that \emph{DXR} is the ancestral node of \emph{CMK}, \emph{MECPS}, and \emph{HDS}, are consistent with the conclusions in \citet{wille2004sparse}.
In the baseline, the nodes with edge connections to \emph{DXR} are: \emph{\{DXPS1, DXPS2, DXPS3, MCT, CMK, MECPS, HDS, UPPS1\}}.
The nodes connected by directed edges pointing to \emph{DXR} are \emph{\{DXPS1, DXPS2, DXPS3\}}, and the node \emph{MCT} is connected by directed edges pointing from \emph{DXR}. Undirected edges connect other nodes.

\textbf{Target=PPDS1.}
Our method gets $\mathit{Pa}(\mathit{PPDS1})=\{\mathit{HDR}\}$, and $\mathit{Ch}(\mathit{PPDS1})$=\emph{\{PPDS2, DPPS2\}}.
We found that the connections among three genes \emph{HDR}, \emph{PPDS2}, and \emph{DPPS2} are consistent with the conclusions in \citet{wille2004sparse}.
In the baseline, the nodes with edge connections to \emph{PPDS1} are: \emph{\{HDR, IPPI1, PPDS2, DPPS2\}}.
And the node \emph{IPPI1} is connected by directed edges pointing from \emph{PPDS1}. Other nodes are connected by undirected edges.

\textbf{Target=MECPS.} 
Our method gets $\mathit{Pa}(\mathit{MECPS})$=\emph{\{DXR, FPPS2\}}, and $\mathit{Ch}(\mathit{MECPS})$=\emph{\{MCT\}}.
We found that the connections among two genes \emph{DXR} and \emph{MCT}, as well as the information that \emph{DXR} is the ancestral node of \emph{MECPS}, are consistent with the conclusions in \citet{wille2004sparse}.
In the baseline, the nodes with edge connections to \emph{MECPS} are: \emph{\{DXR, MCT, CMK, HDS, ACCT1, HMGR2\}}.
The node \emph{HDS} is connected by directed edges pointing to \emph{MECPS}, and the node \emph{CMK} is connected by directed edges pointing from \emph{MECPS}. Undirected edges connect other nodes.

\section{Conclusion and and Further Work}
We introduce a novel local causal discovery algorithm, MMB-by-MMB, designed to be effective in models with the presence of latent variables.
Unlike existing global algorithms, MMB-by-MMB method demonstrates the capability to identify causal structures under equivalent identification conditions, yet it accomplishes this with significantly lower computational expense.
Furthermore, we provide proof validating the correctness of the MMB-by-MMB algorithm.

It should be noted that due to the presence of latent variables, the results of the proposed method still include some instances where it is challenging to determine the causes and effects from purely observational data without any further assumptions. Therefore, exploring how to utilize background knowledge, such as leveraging data generation mechanisms \cite{kaltenpoth2023nonlinear} or expert knowledge \cite{wang2023sound}, to further aid in identifying causes and effects within local structures remains a future research direction. Additionally, leveraging theories on combining interventional and observational data \cite{hauser2015jointly} to learn the local causal structure in the presence of latent variables is interesting for future work.








\section*{Acknowledgements}
This research was supported by the Natural Science Foundation of China (62306019) and the Disciplinary funding of Beijing Technology and Business University (STKY202302). Peng Wu is supported by the National Natural Science Foundation of China (No. 12301370). We appreciate the comments from anonymous reviewers, which greatly helped to improve the paper.

\section*{Impact Statements}
Observational causal discovery is crucial for analyzing systems where experimental manipulation is ethically impermissible. Our research concentrates on learning local causal structures in the presence of latent variables. A significant strength of our work is its practical approach to learning the parents and children of a target variable of interest from finite samples. Our work can be particularly valuable in the realm of complex systems, such as gene and social networks, to improve the interpretability of machine learning systems when deployed in real-world settings.



\nocite{langley00}

\bibliography{example_paper}
\bibliographystyle{icml2024}

\newpage
\appendix
\onecolumn

\section{Notations and Definitions}

\begin{center}
  \begin{table}[htp]
    \centering
    \small
    \begin{tabular}{p{3cm}p{13cm}}
    \toprule
    \textbf{Symbol} & \textbf{Description}\\ 
    \midrule
    $\mathcal{G}$       & A mixed graph  \\
    $\mathcal{M}$       & A Maximal Ancestral Graph (MAG)\\
    $\mathcal{M}_{L}$   & A local Maximal Ancestral Graph (MAG)\\
    $\mathcal{P}$       & A Partial Ancestral Graph (PAG)\\
    $\mathbf{V}$      & The set of all variables \\
    $\mathbf{O}$    & The set of observed variables  \\
    $\mathbf{L}$       & The set of latent variables   \\
    $\mathit{Pa(T)}$,$\mathit{Ch(T)}$  & The set of all parents and children of $T$, respectively \\
    $\mathit{Sp(T)}$ & The set of all spouses of $T$ \\
    $\mathit{An(T)}$, $\mathit{De(T)}$  & The set of all ancestors and descendants of $T$, respectively \\
    $\mathit{Adj(T)}$ & The set of adjacent vertices of $T$\\
    $\mathit{MB(T)}$ & The {{Markov blanket}} of a vertice $T$ in a DAG\\
    $\mathit{MMB(T)}$ & The {{Markov blanket}} of a vertice $T$ in a MAG\\
    $\mathit{MMB^{+}(T)}$  & The set of $\{\mathit{MMB(T)} \cup T\}$\\
    $S_{X,Y}$   & The set of m-separates X and Y\\
    $(\mathbf{X} \CI \mathbf{Y} | \mathbf{Z})_{\mathcal{G}}$ & A set $\mathbf{Z}$ m-separates $\mathbf{X}$ and $\mathbf{Y}$ in $\mathcal{G}$\\
    $(\mathbf{X} \CI \mathbf{Y} | \mathbf{Z})_{P}$ & $\mathbf{X}$ is statistically independent of $\mathbf{Y}$ given $\mathbf{Z}$. We drop the subscript $P$ whenever it is clear from context.\\
    $(\mathbf{X} \nCI \mathbf{Y} | \mathbf{Z})_{P}$ & $\mathbf{X}$ is not statistically independent of $\mathbf{Y}$ given $\mathbf{Z}$\\
    $A \rightarrow B$ in $\mathcal{G}$ & $A$ is a cause of $B$, but $B$ is not a cause of $A$\\
    $A \leftrightarrow B$ in $\mathcal{G}$ & $A$ is not a cause of $B$, and $B$ is not a cause of $A$\\
    $A \quad  B$ in $\mathcal{P}$ & $A$ and $B$ are not adjacent\\
    $A \circ \!\!\rightarrow  B$ in $\mathcal{P}$ & $B$ is not an ancestor of $A$\\
    $A \circ \!\! - \!\! \circ B$ in $\mathcal{P}$ & No set m-separates $A$ and $B$\\
    $A \rightarrow B$ in $\mathcal{P}$ & $A$ is a cause of $B$\\
    $A \leftrightarrow B$ in $\mathcal{P}$ & There is a latent common cause of $A$ and $B$\\
    $\mathcal{L}_{\mathcal{V}}$ & The local structure learned from a subset $\mathcal{V}$ of $\mathbf{V}$, utilizing the test of conditional independence and orientation of V-structures\\
    $\mathit{MMB_{alg}}$  & The algorithm used for learning $\mathit{MMB}$\\
    $\mathbf{WaitList}$ & The list of nodes to be checked by \autoref{theorem1} and \autoref{theorem2}\\ $\mathbf{DoneList}$ & The list of nodes whose local structures have been found\\
    
    \bottomrule
    \end{tabular}
    \caption{The list of main symbols used in this paper}
    \end{table}  
\end{center}

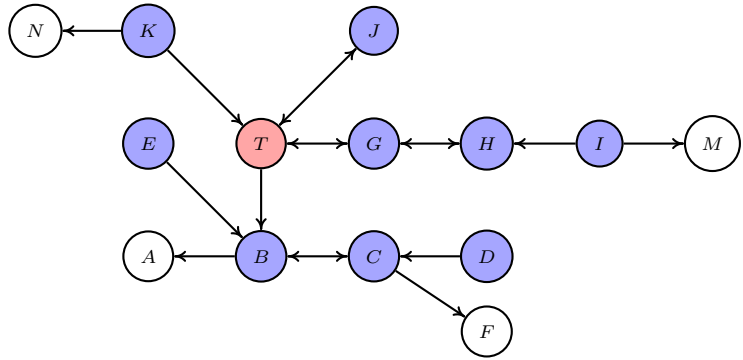
\begin{figure}[h]
    \centering
    \begin{tikzpicture}[node distance=1.5cm and 2cm, thick]
    \draw (4.5, -1.0) node(F) [circle, draw, minimum size=1mm] {{\scriptsize\,$F$\,}};
    \draw (0.0, 0.0) node(A) [circle, draw, minimum size=1mm] {{\scriptsize\,$A$\,}};
    \draw (1.5, 0.0) node(B) [circle, draw, fill=blue!35, minimum size=1mm] {{\scriptsize\,$B$\,}};
    \draw (3.0, 0.0) node(C) [circle, draw, fill=blue!35, minimum size=1mm] {{\scriptsize\,$C$\,}};
    \draw (4.5, 0.0) node(D) [circle, draw, fill=blue!35, minimum size=1mm] {{\scriptsize\,$D$\,}};
    \draw (0.0, 1.5) node(E) [circle, draw, fill=blue!35, minimum size=1mm] {{\scriptsize\,$E$\,}};
    \draw (1.5, 1.5) node(T) [circle, draw, fill=red!35, minimum size=1mm] {{\scriptsize\,$T$\,}};
    \draw (3.0, 1.5) node(G) [circle, draw, fill=blue!35, minimum size=1mm] {{\scriptsize\,$G$\,}};
    \draw (4.5, 1.5) node(H) [circle, draw, fill=blue!35, minimum size=1mm] {{\scriptsize\,$H$\,}};
    \draw (6.0, 1.5) node(I) [circle, draw, fill=blue!35, minimum size=1mm] {{\scriptsize\,$I$\,}};
    \draw (7.5, 1.5) node(M) [circle, draw,  minimum size=1mm] {{\scriptsize\,$M$\,}};
    \draw (-1.5, 3.0) node(N) [circle, draw, minimum size=1mm] {{\scriptsize\,$N$\,}};
    \draw (0.0, 3.0) node(K) [circle, draw, fill=blue!35, minimum size=1mm] {{\scriptsize\,$K$\,}};
    \draw (3.0, 3.0) node(J) [circle, draw, fill=blue!35, minimum size=1mm] {{\scriptsize\,$J$\,}};
    
    \draw[-arcsq] (B) -- (A) node[pos=0.5,sloped,above] {};
    \draw[-arcsq] (E) -- (B) node[pos=0.5,sloped,above] {};
    \draw[double arrow] (B) -- (C) node[pos=0.5,sloped,above] {};
    \draw[-arcsq] (C) -- (F) node[pos=0.5,sloped,above] {};
    \draw[-arcsq] (D) -- (C) node[pos=0.5,sloped,above] {};
    \draw[-arcsq] (T) -- (B) node[pos=0.5,sloped,above] {};
    \draw[-arcsq] (K) -- (T) node[pos=0.5,sloped,above] {};
    \draw[-arcsq] (K) -- (N) node[pos=0.5,sloped,above] {};
    \draw[double arrow] (T) -- (J) node[pos=0.5,sloped,above] {};
    \draw[double arrow] (T) -- (G) node[pos=0.5,sloped,above] {};
    \draw[double arrow] (G) -- (H) node[pos=0.5,sloped,above] {};
    \draw[-arcsq] (I) -- (H) node[pos=0.5,sloped,above] {};
    \draw[-arcsq] (I) -- (M) node[pos=0.5,sloped,above] {};

    \end{tikzpicture}~~~~~~
    \vspace{-3mm}
    \caption{The illustrative example for MMB, where T is the target of interest and the blue nodes belong to $\mathit{MMB(T)}$.}
    \label{Fig-MMB}
\vspace{-3mm}
\end{figure}

\begin{definition}{\textbf{MAG Markov Blanket (MMB)} \cite{pellet2008using}}
In a MAG, the Markov blanket of a vertice $T$, noted as $\mathit{MMB(T)}$, consists of the set of parents, children, children's parents of $\ T$, as well as the district of $\ T$ and of the children of $\ T$, and the parents of each node of these districts, where the district of a node $X$ is the set of all nodes reachable from $X$ using only bidirected edges. 
    \label{definition-MMB}
\end{definition}
In \autoref{Fig-MMB}, the MAG Markov Blanket of $T$ is specifically illustrated.
\section{Discussion of violating the faithfulness assumption}
First, we would like to mention that classical causal discovery, such as PC, FCI, and RFCI, is usually dependent on the causal faithfulness assumption, and these methods have been used in a range of fields \cite{spirtes2016causal}.

Second, reducing unnecessary conditional independence tests can mitigate statistically weak violations of the causal faithfulness assumption \cite{isozaki2014robust}, which is precisely the focus of our paper. 
Moreover, experimental results also validate this point (nTest is minimal, and other metrics are superior to existing methods).

Lastly, incorporating elements of the Greedy Equivalence Search (GES)  algorithm \cite{chickering2002optimal}, a representative score-based method, could be made more robust against violation of faithfulness \cite{zhalama2017weakening}, which is the direction of our future work.

\section{Proofs}\label{Appendix-proofs}

\subsection{Proof of \autoref{theorem1}}
Before presenting the proof, we quote the Theorem 1 of \citet{xie2008recursive}.
\begin{lemma}\label{lemma for theorem1}
Suppose that $A\CI B|C$. Let$~u\in A~and~v\in A\cup C$. Then$~u~$and$~v~$are m-separated by a subset of $A\cup B\cup C$ if and only if they are m-separated by a subset of $A \cup C$.
\end{lemma}
We begin to utilize Lemma 1 to prove Theorem 1.
\begin{proof}
From the property of MMB, we know $T \CI \{\mathbf{O}\setminus MMB^{+}(T)\} | MMB(T)$. 
Let $X$ be a node in $\mathit{MMB(T)}$. According to Lemma \autoref{lemma for theorem1}, we directly obtain that $T$ and $X$ are m-separated by a subset of $\mathbf{O}$ if and only if they are m-separated by a subset of $\mathit{MMB(T)}\setminus \left \{ X \right \}$.
\end{proof}

\subsection{Proof of \autoref{theorem2}}
\begin{proof}
We prove statements $\mathcal{S}1$ and $\mathcal{S}2$ in \autoref{theorem2} separately below. For notational convenience, let $S_{X, Y}$ denote the set of nodes that m-separates $X$ and $Y$.

\textbf{Statement $\mathcal{S}$1.} 
Without loss of generality, we assume that $ V_a  \ast\!\!\rightarrow T \leftarrow\!\!\ast V_b $ is a V-structure in the ground-truth MAG $\mathcal{M}$ over $\mathbf{O}$. 
Because $V_a$ and $V_b$ are two nodes in $\mathit{MMB}(T)$ and according to \autoref{theorem1}, we can ascertain the presence of direct edges $V_a \-- T$ and $ T\-- V_b$ by the marginal distribution of $\mathit{MMB}^{+}(T)$. Next, we need to discuss the following two cases:
\begin{itemize}[leftmargin=15pt,itemsep=0pt,topsep=0pt,parsep=0pt]
    \item \emph{$S_{V_{a},V_{b}} \subset \mathit{MMB}(T)$}. Because $S_{V_{a},V_{b}} \subset \mathit{MMB}(T)$ in the sub-MAG $\mathcal{M}^{\prime}$, we can directly verify the condition $V_a \CI 
 V_b | S_{V_{a},V_{b}}$ from the marginal distribution of $\mathit{MMB}^{+}(T)$. Due to $T \in \mathit{MMB}^{+}(T)$, we obtain $V_a \nCI 
 V_b | S_{V_{a},V_{b}} \cup \{T\}$. This will imply that the $ V_a  \ast\!\!\rightarrow T \leftarrow\!\!\ast V_b $ identified by the marginal distribution of $\mathit{MMB}^{+}(T)$ is exactly the V-structure in the ground-truth MAG $\mathcal{M}$.
    \item \emph{$S_{V_{a},V_{b}} \nsubseteq \mathit{MMB}(T)$}. Because $S_{V_{a},V_{b}} \nsubseteq \mathit{MMB}(T)$ in the sub-MAG $\mathcal{M}^{\prime}$, we can directly deduce that it is impossible to find a separating set for $V_a$ and $V_b$ in $\mathit{MMB}^{+}(T)$, implying that we are unable to identify such V-structures, even if they exist in the ground-truth MAG $\mathcal{M}$.
\end{itemize}

Based on the above analysis, the V-structures we identify from the marginal distribution of $\mathit{MMB}^{+}(T)$ must be consistent with those in the ground-truth MAG $\mathcal{M}$.


\textbf{Statement $\mathcal{S}$2.} Without loss of generality, we assume that $T  \ast\!\!\rightarrow V_a \leftarrow\!\!\ast V_b$ is a V-structure in the ground-truth MAG $\mathcal{M}$ over $\mathbf{O}$. To identify this V-structure from the marginal distribution of $\mathit{MMB^{+}(T)}$, we need to verify the following four conditions: 
\begin{itemize}[leftmargin=14pt,itemsep=0pt,topsep=0pt,parsep=0pt]
      \item 1. $\forall S \subseteq \mathit{MMB(T)}, T \nCI V_a \mid S$ 
      \item 2. $\exists S_{T,V_b} \subseteq \mathit{MMB(T)}, T \CI V_b \mid S_{T,V_b}$ 
      \item 3. $V_a \notin S_{T,V_b}$
      \item 4. $\forall S \subseteq \mathit{MMB^{+}(T)}, V_a \nCI V_b \mid S$
\end{itemize}
According to Theorem \ref{theorem1}, we can directly conclude that there exists the direct edge $T \-- V_a$, and the direct edge $ T\-- V_b$ does not exist, by the marginal distribution of $\mathit{MMB}^{+}(T)$. These results will imply that the above two conditions hold, i.e., $\forall S \subseteq \mathit{MMB(T)}, T \nCI V_a \mid S$  and $\exists S_{T, V_b} \subseteq \mathit{MMB(T)}, T \CI V_b \mid S_{T, V_b}$. Because of the property of V-structure, we obtain the third condition hold, i.e., $V_a \notin S_{T,V_b}$.

We next show that condition 4 can be verified by the marginal distribution of $\mathit{MMB^{+}(T)}$. Let's consider the scenario of a spurious edge between $V_a$ and $V_b$, meaning there is no direct edge between them, but for any subset $\forall S \subseteq \mathit{MMB^{+}(T)}, V_a \nCI V_b \mid S$. Our objective is to demonstrate that if it is an active path connecting $V_a$ and $V_b$ instead of a direct edge, then $\exists S \subseteq \mathit{MMB}^{+}(T), V_a \CI V_b \mid S$. We can examine active paths in the following three cases:
      \begin{itemize}[leftmargin=14pt,itemsep=0pt,topsep=0pt,parsep=0pt]
          \item 1. $V_a \leftarrow \cdots V_b$
          \item 2. $V_a \leftrightarrow V_x\rightarrow\cdots V_b$
          \item 3. $V_a \rightarrow \cdots V_b$
      \end{itemize}      
      Given that $T \ast\!\! \rightarrow V_a$, we have $\mathit{Pa(V_a)},\mathit{Sp(V_a)},\mathit{Pa(\mathit{Sp(V_a)})} \subset \mathit{MMB^{+}(T)}$. Each path in case 1 is blocked by $\mathit{Pa(V_a)}$ and, consequently, by a subset of $\mathit{MMB^{+}(T)}$.
      For case 2, all paths are obstructed by $\mathit{Sp(V_a)}$, which is also a subset of $\mathit{MMB^{+}(T)}$. 
      Thus, we only need to demonstrate that there exists a subset of $\mathit{MMB^{+}(T)}$ that blocks each path $V_a \rightarrow V_x \cdots V_b$. By conditions 2 and 3, $S_{T,V_b}$ blocks each path $V_a \rightarrow V_x \cdots V_b$ since $V_a \notin S_{T,V_b}$.
      In summary, in the absence of a direct edge between $V_a$ and $V_b$, we can obtain that $\exists S\subseteq \mathit{MMB^{+}(T)}, V_a \CI V_b \mid S$. This will imply that condition 4 can be verified by the marginal distribution of $\mathit{MMB}^{+}(T)$.
      
      In conclusion, the V-structure $T  \ast\!\!\rightarrow V_a \leftarrow\!\!\ast V_b$ we identify from the marginal distribution of $\mathit{MMB}^{+}(T)$ must be consistent with those in the ground-truth MAG $\mathcal{M}$.

      
      
\end{proof}

\subsection{Proof of \autoref{theorem3}}
Rule $\mathcal{R}1$ implies that all the causal information of interest, i.e., the edges and directions connected to the target, has been found. Rule $\mathcal{R}2$ asserts that all nodes have been effectively utilized, leaving no node for sequential learning. Both rules $\mathcal{R}1$ and $\mathcal{R}2$ are self-evident.
Therefore, our task is to establish the validity of $\mathcal{R}3$.
Before that, we quote the following lemma since it is used to prove \autoref{theorem3}.
\begin{lemma}\label{Lemma-Joint-probability}
In a MAG $\mathcal{M}$ with a set of vertices $\mathbf{X}$, consider $Y$ as a leaf node (i.e., $Y$ is not an ancestor of any node in $\mathbf{X}$). Let $\mathcal{M}^{\prime}$ be the new MAG obtained by removing $Y$ from $\mathcal{M}$, and $\mathbf{X}^{\prime}$ be the set of all nodes in $\mathcal{M}^{\prime}$, then the following condition holds:
\begin{align}
    P_{\mathcal{M}}(\mathbf{X}^{\prime})=P_{\mathcal{M}^{\prime}}(\mathbf{X}^{\prime})
\end{align}
\end{lemma}
Lemma \ref{Lemma-Joint-probability} implies that the joint probability distribution of the remaining node set $\mathbf{X}^{\prime}$ in the new MAG $\mathcal{M}^{\prime}$ is equivalent to the joint probability distribution of the same node set $\mathbf{X}^{\prime}$ in the original MAG $\mathcal{M}$. In other words, removing the leaf node $Y$ from $\mathcal{M}$ does not alter the joint probability distribution of the remaining node set $\mathbf{X}^{\prime}$.

We now prove Lemma \ref{Lemma-Joint-probability}. 
\begin{proof}  
\begin{equation}
    \begin{aligned}
        P_{\mathcal{M}}(\mathbf{X}^{\prime})& =\sum_{Y}P_{\mathcal{M}}(\mathbf{X}^{\prime},Y)  \\
        &=\sum_Y\prod_{X \in \mathbf{X}^{\prime}}P(X\mid Pa(X))P(Y\mid Pa(Y)) \\
        &=\prod_{X \in \mathbf{X}^{\prime}}P(X\mid Pa(X))\sum_{Y}P(Y\mid Pa(Y)) \\
        &\text{(Because\ }Y\text{ is a leaf node, then }\forall X \in \mathbf{X}^{\prime},Y\notin Pa\left(X\right)) \\
        &=\prod_{X \in \mathbf{X}^{\prime}}P(X\mid Pa(X)) \\
        &=P_{\mathcal{M}^{\prime}}(\mathbf{X}^{\prime})
    \end{aligned}
\end{equation}
\end{proof}

\begin{remark}
    Lemma \autoref{Lemma-Joint-probability} inspires us the following facts:
    \begin{itemize}[leftmargin=14pt,itemsep=0pt,topsep=0pt,parsep=0pt]
        \item Firstly, let $\mathbf{O}$ be a set of observed data, $\mathcal{M}$ be the MAG graph over $\mathbf{O}$, and  $\mathbf{X}^{\prime}=\{\mathbf{O}\setminus \mathbf{Y}\}$. $\mathbf{Y}$ represents the set of leaf nodes relative to $\mathbf{X}^{\prime}$ in $\mathcal{M}$. According to Lemma \ref{Lemma-Joint-probability}, we can deduce that $P_{\mathcal{M}}(\mathbf{X}^{\prime})=P_{\mathcal{M}^{\prime}}(\mathbf{X}^{\prime})$, where $\mathcal{M}^{\prime}$ is the new MAG obtained by removing $\mathbf{Y}$ from $\mathcal{M}$.
        \item Secondly, let $\mathbf{X}^{\prime\prime}=\{\mathbf{O}\setminus \mathbf{Y} \cup \mathbf{Y}^{\prime}\}$, and $\mathbf{Y}^{\prime}$ denote the set of leaf nodes relative to $\mathbf{X}^{\prime\prime}$ in $\mathcal{M}^{\prime}$. Subsequently, we can infer $P_{\mathcal{M}^{\prime}}(\mathbf{X}^{\prime\prime})=P_{\mathcal{M}^{\prime\prime}}(\mathbf{X}^{\prime\prime})$ using Lemma \ref{Lemma-Joint-probability}, where $\mathcal{M}^{\prime\prime}$ is the new MAG obtained by removing $\mathbf{Y}^{\prime}$ from $\mathcal{M}^{\prime}$.
        \item Finally, by repeating the above steps multiple times, we can obtain a local MAG $\mathcal{M}_{L}$, and $P_{\mathcal{M}}(\mathbf{X})=P_{\mathcal{M}^{\prime}}(\mathbf{X})=P_{\mathcal{M}^{\prime\prime}}(\mathbf{X})=P_{\mathcal{M}_{L}}(\mathbf{X})$, where $\mathbf{X}=\{\mathbf{O}\setminus \mathcal{Y} \}$, and $\mathcal{Y}$ denotes the leaf nodes that are removed during the repetition process.
    \end{itemize}
\label{Remark-Joint-Distribution}
\end{remark}

Based on the above analysis, we now prove the rule $\mathcal{R}3$ of \autoref{theorem3}.
\begin{proof}
    Based on the above description of Remark \ref{Remark-Joint-Distribution} and relying on the faithfulness assumption, it can be inferred that after iteratively deleting all leaf nodes in the MAG, the relationship among the remaining nodes will remain unchanged.
    
    First, let's provide a more detailed explanation of $\mathcal{R}3$. In our sequential approach, assuming the sequential learning process terminates due to satisfying $\mathcal{R}3$, the result we learn is a local MAG's PAG $\mathcal{P}$. Assuming that we have identified a node set $\mathbf{O}^{\prime}$ around $T$ in the sequential learning process, and there exists a path between each node in the set $\mathbf{Waitlist}$ and $T$. Suppose $Y \in \mathbf{Waitlist}$, let $V_i$ represent the nodes excluding $T$ and $Y$ in the path, where $i \in [1, n]$ and $V_i \in \mathbf{Donelist}$. Let $V_1$ denote the node closest to $T$ on the path, and $V_n$ denote the node closest to $Y$. If we identify that all paths connected by undirected edges around $T$ possess the following characteristics:
    the edges between $T$ and $V_1$ on the path are undirected, while the edges between $V_n$ and $Y$ are either $V_n\rightarrow Y$ or $V_n\circ\!\!\rightarrow Y$. Hence, upon satisfaction of $\mathcal{R}3$, we can conclude the sequential learning algorithm.

    Then, We proceed to demonstrate why the sequential learning algorithm can be halted when $\mathcal{R}3$ is satisfied. In our learning process, we identify that the edges between $T$ and $V_1$ on the paths are undirected, while $Y$ are not ancestors of $V_n$. 
    These paths from $T$ to $Y$ in underlying $\mathcal{M}_{L}$ can be considered in the following two cases:
    \begin{itemize}[leftmargin=14pt,itemsep=0pt,topsep=0pt,parsep=0pt]
        \item 1. $T$ $\cdots V_1 \cdots  V_n\leftrightarrow$ $Y$\\
        \item 2. $T$ $\cdots V_1 \cdots  V_n\rightarrow$ $Y$    
    \end{itemize}
    Since we have identified $\ast\!\!\rightarrow Y$, we can infer that these $Y$ nodes belong to the leaf nodes of the underlying $\mathcal{M}_{L}$. Combining these $Y$ nodes into a set $\mathbf{Y}$, according to Lemma \autoref{Lemma-Joint-probability}, we can deduce that $P_{\mathcal{M}_{L}^{\prime}}(\mathbf{O}^{\prime\prime})=P_{\mathcal{M}_{L}}(\mathbf{O}^{\prime\prime})$, where $\mathbf{O}^{\prime\prime} = \{ \mathbf{O}^{\prime}\setminus \mathbf{Y}\}$ is the new MAG obtained by removing $\mathbf{Y}$ from $\mathcal{M}_{L}$.
    
    This implies that the joint probability distribution of the remaining nodes set $\mathbf{O}^{\prime\prime}$ in $\mathcal{M}_{L}^{\prime}$ is equivalent to the joint probability distribution of the same node set $\mathbf{O}^{\prime\prime}$ in the $\mathcal{M}_{L}$. Then, through $P_{\mathcal{M}}(\mathbf{X})=P_{\mathcal{M}^{\prime}}(\mathbf{X})=P_{\mathcal{M}^{\prime\prime}}(\mathbf{X})=P_{\mathcal{M}_{L}}(\mathbf{X})$ in \autoref{Remark-Joint-Distribution}, we can get $P_{\mathcal{M}_{L}^{\prime}}(\mathbf{O}^{\prime\prime})=P_{\mathcal{M}}(\mathbf{O}^{\prime\prime})$.
    
    However, we failed to identify, based on the marginal distribution of $\mathbf{O}^{\prime\prime}$, that all paths involving undirected edges connected to $T$ are blocked by $\ast\!\!\rightarrow$. Therefore, we continue the learning process until all paths involving undirected edges connected to $T$ are blocked by $\ast\!\!\rightarrow$ through the marginal distribution of $\mathbf{O}^{\prime}$.
    To summarize, when the situation satisfying $\mathcal{R}3$ is identified,
    we can get $P_{\mathcal{M}_{L}}(\mathbf{O}^{\prime})=P_{\mathcal{M}}(\mathbf{O}^{\prime})$
    which implies that continuing this algorithm will not contribute to orienting the undirected edges in $\mathcal{P}$. 
    Hence, upon satisfaction of $\mathcal{R}3$, we can conclude the sequential learning algorithm. 
\end{proof}

\subsection{Proof of \autoref{theorem-4}}
\begin{proof}
To establish the correctness of the MMB-by-MMB approach, it is imperative to demonstrate the correctness of all edges and orientations in the resulting graph $\mathcal{P}$. Additionally, it is crucial to assert that the undirected edges linked to the target node $T$ remain unaltered, defying further orientation even as the algorithm progresses.
    
    Following \autoref{theorem1}, it is established that all edges connected to nodes in the Donelist are accurate. Given that $T$ is encompassed within the Donelist, the edges linked to $T$ are deemed correct.
    
    Subsequently, relying on \autoref{theorem2}, it can be inferred that all v-structures in $\mathcal{P}$ are correct, and those v-structures having at least one node that does not belong to the ancestors of the collider within the Donelist are correctly identified. Following Zhang's orientation methodology, the undirected edges in $\mathcal{P}$ are oriented by checking the presence of edges in $\mathcal{P}$.
    
    Ultimately, we demonstrate that continuing the algorithm cannot orient the undirected edges connected to $T$ in the output $\mathcal{P}$ if they are present. As outlined earlier, we accurately ascertain all edges and v-structures along with their orientations. Thus, a PAG $\mathcal{P}$ representing the Markov equivalence class of the underlying MAG is obtained when the Donelist equals the complete set $\mathbf{O}$ of all nodes. In cases where the Donelist is a subset of $\mathbf{O}$, the algorithm stops, as nodes in the Donelist do not establish connections with nodes outside $\mathcal{P}$, or their paths to nodes outside $\mathcal{P}$ are all blocked by $\ast \rightarrow$. In such instances, the edges and orientations identified through the continuing algorithm do not contribute to orienting the undirected edges in $\mathcal{P}$, as these undirected edges have already been enveloped by previously determined directed edges.
    
    Hence, the correctness of the MMB-by-MMB algorithm is proven.
\end{proof}

\section{Illustration of MMB-by-MMB Algorithm}\label{Appendix-Algorithm}
In this section, we illustrate our MMB-by-MMB with the graph in \autoref{fig:inferred local PAG process}(a).
We assume Oracle tests for conditional independence conditions.
\begin{itemize}[leftmargin=14pt,itemsep=0pt,topsep=0pt,parsep=0pt]
    \item It first initializes sets $\mathbf{Waitlist}$ = $\{ V_5\}$, $\mathbf{Donelist}$ = $\emptyset$, and graph $\mathcal{P}=\emptyset$ (Line 1).
    \item After initialization, it runs $\mathit{MMB_{alg}(V_5)}$ and obtain $\mathit{MMB(V_5)}\!\!=\!\!\{V_3, V_4, V_7, V_8, V_{10}, V_{12}\}$ (Lines $3 \sim 4$). 
    \item It then learns $\mathit{L_{MMB^{+}(V_5)}}$ over $\mathit{MMB^{+}(V_5)}$: $V_4 \circ\!\!\rightarrow V_5\circ\!\!\rightarrow V_{10} \leftarrow\!\!\circ V_3 \circ\!\!\rightarrow V_4$ and $V_{12}\circ\!\!\rightarrow V_8 \leftrightarrow V_5 \leftarrow\!\!\circ V_4 \leftarrow\!\!\circ V_7 \circ\!\!\rightarrow V_8$,  as depicted in \autoref{fig:inferred local PAG process}(b) (Line 10).
    \item Next, it updates $\mathcal{P}$ by selecting the edges connected to $V_X$ and the V-structures containing $V_X$ (Line 12). According to \autoref{theorem1} and \autoref{theorem2}, these edges can be determined :$V_3 \circ\!\!\rightarrow V_{10} \leftarrow\!\!\circ V_5 \leftarrow\!\!\circ V_4$, $V_5 \leftrightarrow V_8 \leftarrow\!\!\circ V_{12}$ and $V_7 \circ\!\!\rightarrow V_8$.
    \item It now orients $V_5 \circ\!\!\rightarrow V_{10}$ as $V_5 \rightarrow V_{10}$ by orientation rules (Line 13). Consequently, it obtains $\mathcal{P}$ as shown in \autoref{fig:inferred local PAG process}(c).
    \item Then, it updates $\mathbf{Donelist}\!\!=\!\!\{V_5\}$, and $\mathbf{Waitlist}\!\!=\!\!\{V_4, V_8, V_{10}\}$ (Lines $14 \sim 15$).
    \item Sequentially, it runs $\mathit{MMB_{alg}(V_4)}$ and obtain $\mathit{MMB}(V_4)\!\!=\!\!\{V_2,V_3,V_5,V_7,V_8,V_{11},V_{12}\}$.
    \item It then learns the local structure $\mathcal{L}_{\mathit{MMB^{+}(V_4)}}:V_{11} \circ\!\!\rightarrow V_3 \leftrightarrow V_4 \circ\!\!\rightarrow V_5 \leftrightarrow V_8$ and $V_4 \leftarrow\!\!\circ V_2 \circ\!\!\--\!\!\circ V_7 \circ\!\!\rightarrow V_8 \leftarrow\!\!\circ V_{12}$,  as shown in \autoref{fig:inferred local PAG process} (Line 10).
    \item According to \autoref{theorem1} and \autoref{theorem2}, these edges can be determined : $V_{11} \circ\!\!\rightarrow V_3 \leftrightarrow V_4 \circ\!\!\rightarrow V_5 \leftrightarrow V_8 \leftarrow\!\!\circ V_{12}$, $V_2 \circ\!\!\rightarrow V_4$ and $V_{12} \circ\!\!\rightarrow V_8$. Next, it pools the determined edges together and orients $V_4 \circ\!\!\rightarrow V_5$ as $V_4 \rightarrow V_5$ (Lines $12 \sim 13$). Following this, it derives the local structure $\mathcal{P}$, as illustrated in \autoref{fig:inferred local PAG process}(e).
    \item Next, it updates sets $\mathbf{Waitlist}\!\!=\!\!\{V_8,V_{10},V_2,V_3\}$, $\mathbf{Donelist}\!\!=\!\!\{V_5, V_4\}$ (Lines $14 \sim 15$).
    \item Finally, the algorithm terminates because stop $\mathcal{R}1$ is satisfied. Output the resulting local structure $\mathcal{P}$, which is depicted in \autoref{fig:inferred local PAG process}(f).
\end{itemize}
The ultimate local $\mathcal{P}$, acquired through orienting rules, is presented in \autoref{fig:inferred local PAG process}.(f). As all edges connected to the target $V_5$ have been oriented (i.e., satisfying stop $\mathcal{R}1$), the learning process can be concluded.

\section{More Results on Experiments}\label{Appendix-simulations}
All experiments were performed with Intel 2.90GHz and 2.89 GHz CPUs and 128 GB of memory. We give more experimental results here.

Table \ref{tab:datasets} provides a detailed overview of the network statistics used in this paper.
\begin{table}[H]
	\small
	\center \caption{Statistics on the Networks.}
	\label{tab:datasets}
	\begin{tabular}{|c|c|c|c|}
		\hline Networks & Num.Variables & Avg degree & Max in-degree\\
		\hline \emph{MILDEW} & 35 & 2.63 & 3\\
		\hline \emph{ALARM} & 37 & 2.49 & 4\\
		\hline \emph{WIN95PTS} & 76 & 2.95 & 7\\
		\hline \emph{ANDES} & 223 & 3.03 & 6\\
		\hline 
	\end{tabular}
\end{table}

Tables \ref{table-5} $\sim$ \ref{table-8} provide the complete results in Section \ref{Subsection-Synthetic-Data}.

\begin{table*}[hpt!]
\centering
\small
\caption{Performance Comparisons on MILDEW.Net}
\resizebox{0.98\textwidth}{!}{
\begin{tabular}{cc|ccccc|ccccc}
\toprule
\multicolumn{2}{c}{} & \multicolumn{5}{|c|}{Size=1000} & \multicolumn{5}{c}{Size=5000}\\
\hline
\multicolumn{1}{c}{Target} & Algorithm & {\textbf{Precision}$\uparrow$} & {\textbf{Recall}$\uparrow$} &{\textbf{F1}$\uparrow$} & {\textbf{Distance}$\downarrow$} & {\textbf{nTest}$\downarrow$} & {\textbf{Precision}}$\uparrow$ & {\textbf{Recall}$\uparrow$} &{\textbf{F1}$\uparrow$} & {\textbf{Distance}$\downarrow$} & {\textbf{nTest}$\downarrow$}\\
\hline
& PC-stable     
& 0.28±0.09   & 0.28±0.09   & 0.28±0.09   & 1.02±0.13   & 5066.55    
& 0.27±0.07   & 0.27±0.07   & 0.27±0.07   & 1.04±0.10   & 8917.40    \\
& FCI         
& 0.70±0.25   & 0.64±0.29   & 0.66±0.28   & 0.48±0.39   & 10307.75 
& 0.77±0.25   & 0.72±0.29   & 0.73±0.28   & 0.38±0.39   & 16663.83   \\
& RFCI        
& 0.69±0.26   & 0.62±0.29   & 0.64±0.28   & 0.51±0.39   & 5066.55  
& 0.73±0.25   & 0.68±0.28   & 0.70±0.27   & 0.43±0.38   & 8917.40     \\
{dm-2}& MB-by-MB    
& 0.55±0.11 & 0.62±0.14 & 0.57±0.12 & 0.61±0.17 & 6085.94
& 0.55±0.11 & 0.63±0.15 & 0.57±0.11 & 0.60±0.16 & 21849.32 \\
& CMB         
& 0.55±0.13   & 0.58±0.14   & 0.56±0.13   & 0.62±0.18   & 2169.34   
& 0.56±0.12   & 0.60±0.13   & 0.57±0.12   & 0.60±0.17   & 2573.37   \\
& GraN-LCS    
& 0.68±0.17 & 0.72±0.18 & 0.68±0.16 & 0.46±0.22 & -  
& 0.69±0.18 & 0.72±0.19 & 0.68±0.17 & 0.46±0.24 & -       \\
& MMB-by-MMB  
& \textbf{0.94±0.15}   & \textbf{0.93±0.17}   & \textbf{0.93±0.17}  & \textbf{0.10±0.24}   & \textbf{732.89}    
& \textbf{0.98±0.08}   & \textbf{0.97±0.11 }  & \textbf{0.97±0.10}   & \textbf{0.04±0.15}   & \textbf{1064.33}   \\
\hline
& PC-stable    
& 0.28±0.08 & 0.28±0.09 & 0.28±0.08 & 1.02±0.12 & 5032.70 
& 0.27±0.07 & 0.27±0.07 & 0.27±0.07 & 1.04±0.10 & 8930.25   \\
& FCI        
& 0.83±0.27 & 0.83±0.27 & 0.83±0.27 & 0.24±0.39 & 10260.86
& 0.85±0.27 & 0.85±0.26 & 0.85±0.27 & 0.21±0.38 & 16637.09  \\
& RFCI       
& 0.81±0.28 & 0.81±0.28 & 0.81±0.28 & 0.27±0.39 & 5032.70  
& 0.84±0.26 & 0.84±0.27 & 0.84±0.27 & 0.22±0.38 & 8930.25   \\
{dm-1}& MB-by-MB   
& 0.51±0.13 & 0.57±0.18 & 0.52±0.14 & 0.67±0.20 & 4596.90 
& 0.50±0.12 & 0.58±0.18 & 0.51±0.13 & 0.68±0.19 & 15864.89 \\
& CMB        
& 0.49±0.15 & 0.52±0.17 & 0.50±0.15 & 0.71±0.22 & 3440.64  
& 0.50±0.13 & 0.54±0.16 & 0.51±0.14 & 0.69±0.20 & 3661.29  \\
& GraN-LCS  
& 0.70±0.22 & 0.78±0.20 & 0.73±0.22 & 0.39±0.30 & -
& 0.67±0.21 & 0.74±0.20 & 0.69±0.20 & 0.43±0.29 & -                \\
& MMB-by-MMB 
& \textbf{0.95±0.15} & \textbf{0.95±0.15} & \textbf{0.95±0.15} & \textbf{0.07±0.22} & \textbf{392.95}    
& \textbf{0.97±0.13} & \textbf{0.97±0.13} & \textbf{0.97±0.13} & \textbf{0.04±0.18} & \textbf{613.61}   \\
\hline
& Pc-stable   
& 0.25±0.04   & 0.25±0.04   & 0.25±0.04   & 1.06±0.06   & 4983.90    
& 0.25±0.04   & 0.25±0.04   & 0.25±0.04   & 1.06±0.06   & 8852.61    \\
& FCI        
& 0.71±0.27   & 0.70±0.27   & 0.70±0.27   & 0.43±0.39   & 10187.20 
& 0.77±0.30   & 0.77±0.30   & 0.77±0.30   & 0.33±0.43   & 16496.33   \\
& RFCI       
& 0.62±0.28   & 0.61±0.28   & 0.61±0.28   & 0.56±0.40   & 4983.90    
& 0.74±0.30   & 0.74±0.30   & 0.74±0.30   & 0.37±0.43   & 8852.61    \\
{foto-4}& MB-by-MB   
& 0.29±0.07 & 0.37±0.14 & 0.30±0.08 & 0.97±0.13 & 6715.11 
& 0.27±0.04 & 0.34±0.13 & 0.28±0.05 & 1.00±0.09 & 23781.90 \\
& CMB        
& 0.30±0.09   & 0.35±0.13   & 0.31±0.09   & 0.97±0.13   & 2335.02   
& 0.29±0.07   & 0.34±0.12   & 0.30±0.08   & 0.99±0.11   & 3097.41    \\
& GraN-LCS   
& 0.32±0.12 & 0.40±0.18 & 0.34±0.13 & 0.92±0.19 & - 
& 0.33±0.12 & 0.40±0.18 & 0.34±0.14 & 0.92±0.19 & -                  \\
& MMB-by-MMB 
& \textbf{0.93±0.12}   & \textbf{0.92±0.14}   & \textbf{0.92±0.14}   & \textbf{0.11±0.20}   & \textbf{820.24}     
& \textbf{0.95±0.15}   & \textbf{0.95±0.15}   & \textbf{0.95±0.15}   & \textbf{0.07±0.22}   & \textbf{1207.26}     \\
\hline
& Pc-stable   
& 0.33±0.18   & 0.33±0.18   & 0.33±0.18   & 0.94±0.26   & 5071.66    
& 0.33±0.18   & 0.33±0.18   & 0.33±0.18   & 0.95±0.25   & 8972.31     \\
& FCI        
& 0.77±0.20   & 0.70±0.20   & 0.72±0.20   & 0.39±0.29   & 10338.28 
& 0.84±0.15   & 0.77±0.20   & 0.80±0.18   & 0.29±0.26   & 16762.44    \\
& RFCI       
& 0.72±0.21   & 0.64±0.19   & 0.67±0.20   & 0.47±0.28   & 5071.66    
& 0.81±0.15   & 0.73±0.20   & 0.76±0.18   & 0.35±0.26   & 8972.31    \\
{dm-4}& MB-by-MB   
& 0.59±0.14 & 0.69±0.18 & 0.62±0.14 & 0.53±0.20 & 7075.49
& 0.60±0.15 & 0.71±0.18 & 0.63±0.16 & 0.51±0.22 & 28815.65 \\
& CMB        
& 0.60±0.15   & 0.63±0.16   & 0.61±0.15   & 0.55±0.22   & 2325.96  
& 0.58±0.15   & 0.63±0.15   & 0.60±0.15   & 0.57±0.21   & 3638.17    \\
& GraN-LCS   
& 0.57±0.17 & 0.59±0.19 & 0.56±0.16 & 0.62±0.23 & - 
& 0.60±0.20 & 0.61±0.21 & 0.59±0.19 & 0.58±0.26 & -  \\
& MMB-by-MMB 
& \textbf{0.95±0.13}   & \textbf{0.91±0.16}   & \textbf{0.92±0.15}   & \textbf{0.11±0.21}   & \textbf{527.14}    
& \textbf{0.99±0.05}   & \textbf{0.98±0.09}   & \textbf{0.98±0.08}   & \textbf{0.03±0.11}   & \textbf{690.57}  \\
\bottomrule
\end{tabular}
}
\label{table-5}
\begin{tablenotes}
		\item \scriptsize{Note: The symbol '-' indicates that GraN-LCS does not output this information. MMB-by-MMB with the second best result is underlined. $\uparrow$ means a higher value is better, and vice versa.}
\end{tablenotes}
\end{table*}

\begin{table*}[hpt!]
\centering
\small
\caption{Performance Comparisons on ALARM.Net}
\resizebox{0.98\textwidth}{!}{
\begin{tabular}{cc|ccccc|ccccc}
\toprule
\multicolumn{2}{c}{} & \multicolumn{5}{|c|}{Size=1000} & \multicolumn{5}{c}{Size=5000}\\
\hline
\multicolumn{1}{c}{Target} & Algorithm & {\textbf{Precision}$\uparrow$} & {\textbf{Recall}$\uparrow$} &{\textbf{F1}$\uparrow$} & {\textbf{Distance}$\downarrow$} & {\textbf{nTest}$\downarrow$} & {\textbf{Precision}}$\uparrow$ & {\textbf{Recall}$\uparrow$} &{\textbf{F1}$\uparrow$} & {\textbf{Distance}$\downarrow$} & {\textbf{nTest}$\downarrow$}\\
\hline
& PC-stable 
& 0.56±0.16 & 0.55±0.14 & 0.56±0.15 & 0.63±0.20 &3515.85  
& 0.56±0.16 & 0.55±0.14 & 0.56±0.15 & 0.63±0.20 &4878.73  \\
& FCI 
& 0.78±0.25 & 0.77±0.25 & 0.77±0.25 & 0.32±0.35 & 7552.66 
& 0.84±0.24 & 0.83±0.24 & 0.83±0.24 & 0.24±0.34 & 10173.30 \\
& RFCI 
& 0.78±0.25 & 0.77±0.24 & 0.77±0.24 & 0.32±0.34 & 3515.85  
& 0.83±0.23 & 0.83±0.24 & 0.83±0.24 & 0.24±0.33 & 4878.73   \\
{LVEDVOLUME} & MB-by-MB 
& 0.46±0.18 & 0.46±0.18 & 0.45±0.17 & 0.77±0.25 & 1531.45
& 0.44±0.15 & 0.44±0.16 & 0.43±0.14 & 0.81±0.20 & 4196.87 \\
& CMB 
& 0.44±0.21 & 0.43±0.19 & 0.43±0.20 & 0.80±0.28 & 1471.71 
& 0.44±0.20 & 0.43±0.18 & 0.43±0.18 & 0.81±0.26 & 1992.32 \\
& GraN-LCS 
& 0.58±0.15 & 0.57±0.14 & 0.57±0.14 & 0.61±0.20 & -     
& 0.58±0.13 & 0.58±0.13 & 0.58±0.13 & 0.60±0.18 & -  \\
&MMB-by-MMB 
& \textbf{0.97±0.12} & \textbf{0.96±0.12} & \textbf{0.96±0.12} & \textbf{0.05±0.16} & \textbf{324.09} 
& \textbf{0.98±0.09} & \textbf{0.98±0.09} & \textbf{0.98±0.09} & \textbf{0.03±0.12} & \textbf{344.51} \\
\hline
& PC-stable 
& 0.58±0.18 & 0.56±0.15 & 0.57±0.16 & 0.61±0.22 & 3529.88  
& 0.58±0.18 & 0.57±0.15 & 0.57±0.16 & 0.61±0.22 & 4846.68  \\
& FCI 
& 0.76±0.27 & 0.75±0.26 & 0.74±0.26 & 0.36±0.37 & 7629.62 
& 0.80±0.25 & 0.78±0.26 & 0.79±0.26 & 0.30±0.36 & 10199.60  \\
& RFCI 
& 0.74±0.27 & 0.73±0.27 & 0.73±0.27 & 0.39±0.38 & 3529.88  
& 0.76±0.26 & 0.74±0.27 & 0.74±0.26 & 0.37±0.37 & 4846.68 \\
{VENTTUBE} & MB-by-MB 
& 0.36±0.12 & 0.37±0.14 & 0.36±0.12 & 0.90±0.17 & 3081.04 
& 0.32±0.11 & 0.32±0.11 & 0.32±0.11 & 0.97±0.15 & 8077.43 \\
& CMB 
& 0.32±0.13 & 0.31±0.11 & 0.31±0.12 & 0.97±0.17 & 3537.93 
& 0.29±0.10 & 0.29±0.10 & 0.29±0.09 & 1.00±0.13 & 4630.07 \\
& GraN-LCS 
& 0.43±0.13 & 0.46±0.16 & 0.44±0.14 & 0.80±0.20 & -     
& 0.44±0.13 & 0.47±0.15 & 0.45±0.13 & 0.79±0.18 & -       \\
&MMB-by-MMB 
& \textbf{0.87±0.24} & \textbf{0.85±0.24} & \textbf{0.85±0.24} & \textbf{0.21±0.34} & \textbf{570.23}  
& \textbf{0.89±0.21} & \textbf{0.87±0.22} & \textbf{0.87±0.22} & \textbf{0.18±0.31} & \textbf{736.22} \\
\hline
& PC-stable 
& 0.24±0.05 & 0.24±0.05 & 0.24±0.05 & 1.07±0.07 & 3482.37 
& 0.24±0.05 & 0.24±0.05 & 0.24±0.05 & 1.07±0.07 & 4758.09  \\
& FCI 
& 0.63±0.22 & 0.52±0.18 & 0.55±0.19 & 0.64±0.27 & 7565.43
& 0.76±0.17 & 0.66±0.19 & 0.69±0.17 & 0.44±0.25 & 10110.46  \\
& RFCI 
& 0.59±0.20 & 0.47±0.16 & 0.51±0.17 & 0.70±0.23 & 3482.37  
& 0.73±0.17 & 0.63±0.18 & 0.66±0.17 & 0.49±0.24 & 4758.09  \\
{CATECHOL} & MB-by-MB 
& 0.31±0.09 & 0.38±0.13 & 0.33±0.10 & 0.94±0.14 & 6336.96 
& 0.30±0.08 & 0.38±0.12 & 0.32±0.09 & 0.95±0.13 & 20601.93 \\
& CMB 
& 0.29±0.07 & 0.35±0.11 & 0.30±0.08 & 0.97±0.11 & 5036.08
& 0.30±0.07 & 0.36±0.10 & 0.32±0.08 & 0.96±0.11 & 4204.28 \\
& GraN-LCS 
& 0.26±0.07 & 0.26±0.09 & 0.26±0.08 & 1.05±0.11 & -     
& 0.25±0.07 & 0.25±0.06 & 0.25±0.06 & 1.06±0.09 & - \\
&MMB-by-MMB 
& \textbf{0.92±0.15} & \textbf{0.87±0.17} & \textbf{0.88±0.16} & \textbf{0.17±0.23} & \textbf{936.64} 
& \textbf{0.95±0.11} & \textbf{0.94±0.13} & \textbf{0.94±0.13} & \textbf{0.09±0.18} & \textbf{1128.77} \\
\hline
& PC-stable 
& 0.57±0.17 & 0.56±0.13 & 0.56±0.14 & 0.62±0.20 & 3470.71
& 0.57±0.17 & 0.56±0.13 & 0.56±0.14 & 0.62±0.20 & 4840.23\\
& FCI 
& 0.78±0.25 & 0.77±0.25 & 0.77±0.25 & 0.33±0.35 & 7483.00 
& 0.84±0.23 & 0.84±0.23 & 0.84±0.23 & 0.23±0.33 & 10171.28  \\
& RFCI 
& 0.72±0.25 & 0.71±0.24 & 0.71±0.25 & 0.41±0.35 & 3470.71  
& 0.83±0.24 & 0.83±0.24 & 0.83±0.24 & 0.24±0.34 & 4840.23 \\
{STROKEVOLUME} & MB-by-MB 
& 0.43±0.18 & 0.48±0.21 & 0.43±0.18 & 0.79±0.25 & 2076.76
& 0.37±0.15 & 0.41±0.17 & 0.38±0.15 & 0.88±0.21 & 6377.99\\
& CMB 
& 0.44±0.20 & 0.45±0.20 & 0.44±0.19 & 0.79±0.27 & 2197.08
& 0.37±0.19 & 0.37±0.19 & 0.37±0.19 & 0.89±0.26 & 2597.72\\
& GraN-LCS 
& 0.42±0.13 & 0.46±0.16 & 0.43±0.14 & 0.80±0.20 & -     
& 0.42±0.13 & 0.46±0.16 & 0.43±0.14 & 0.80±0.20 & -     \\
&MMB-by-MMB 
& \textbf{0.95±0.15} & \textbf{0.95±0.16} & \textbf{0.95±0.16} & \textbf{0.08±0.22} & \textbf{566.39} 
& \textbf{0.98±0.09} & \textbf{0.98±0.09} & \textbf{0.98±0.09} & \textbf{0.02±0.12} & \textbf{698.17} \\
\bottomrule
\end{tabular}
}
\label{table-6}
\begin{tablenotes}
	\item \scriptsize{Note: The symbol '-' indicates that GraN-LCS does not output this information. MMB-by-MMB with the second best result is underlined. $\uparrow$ means a higher value is better, and vice versa.}
\end{tablenotes}
\end{table*}

\begin{table*}[hpt!]
\centering
\small
\caption{Performance Comparisons on WIN95PTS.Net}
\resizebox{0.98\textwidth}{!}{
\begin{tabular}{cc|ccccc|ccccc}
\toprule
\multicolumn{2}{c}{} & \multicolumn{5}{|c|}{Size=1000} & \multicolumn{5}{c}{Size=5000}\\
\hline
\multicolumn{1}{c}{Target} & Algorithm & {\textbf{Precision}$\uparrow$} & {\textbf{Recall}$\uparrow$} &{\textbf{F1}$\uparrow$} & {\textbf{Distance}$\downarrow$} & {\textbf{nTest}$\downarrow$} & {\textbf{Precision}}$\uparrow$ & {\textbf{Recall}$\uparrow$} &{\textbf{F1}$\uparrow$} & {\textbf{Distance}$\downarrow$} & {\textbf{nTest}$\downarrow$}\\
\hline
&Pc-stable           
& 0.52±0.07 & 0.52±0.07 & 0.52±0.07 & 0.68±0.10 & 12657.62    
& 0.53±0.07 & 0.53±0.08 & 0.53±0.07 & 0.67±0.10 & 26398.40 \\
&FCI                
& 0.69±0.24 & 0.69±0.24 & 0.68±0.24 & 0.45±0.34 & 25417.23  
& 0.77±0.27 & 0.76±0.27 & 0.76±0.27 & 0.34±0.38 & 43850.55 \\
&RFCI               
& 0.67±0.23 & 0.66±0.23 & 0.66±0.23 & 0.48±0.33 & 12657.62    
& 0.77±0.27 & 0.75±0.28 & 0.76±0.28 & 0.35±0.39 & 26398.40 \\
{Problem5}&MB-by-MB           
& 0.46±0.19 & 0.50±0.22 & 0.47±0.20 & 0.75±0.28 & 13633.52 
&NA &NA &NA &NA &NA   \\
&CMB                
& 0.57±0.16 & 0.60±0.18 & 0.58±0.17 & 0.59±0.23  & 4757.95    
& 0.54±0.15 & 0.58±0.17 & 0.56±0.16 & 0.63±0.22 & \textbf{5413.78}   \\
&GraN-LCS           
& 0.48±0.14 & 0.50±0.15 & 0.48±0.14 & 0.73±0.20 & -         
& 0.48±0.16 & 0.48±0.17 & 0.48±0.16 & 0.74±0.22 & -      \\
&MMB-by-MMB 
& \textbf{0.90±0.20} & \textbf{0.90±0.20} & \textbf{0.89±0.20} & \textbf{0.15±0.29} & \textbf{3907.42}     
& \textbf{0.93±0.18} & \textbf{0.92±0.19} & \textbf{0.92±0.19} & \textbf{0.11±0.27} & \underline{12372.20} \\
\hline
& PC-stable 
& 0.78±0.08 & 0.77±0.06 & 0.78±0.07 & 0.32±0.10 & 12637.44 
& 0.77±0.06 & 0.76±0.04 & 0.76±0.05 & 0.34±0.07 & 25058.00  \\
& FCI 
& 0.80±0.24 & 0.80±0.24 & 0.80±0.24 & 0.28±0.34 & 25651.52 
& 0.83±0.24 & 0.83±0.25 & 0.83±0.25 & 0.24±0.35 & 42604.71  \\
& RFCI 
& 0.81±0.24 & 0.81±0.24 & 0.81±0.24 & 0.27±0.34 & 12637.44 
& 0.82±0.25 & 0.82±0.25 & 0.82±0.25 & 0.25±0.35 & 25058.00  \\
{HrglssDrtnAftrPrnt} & MB-by-MB 
& 0.46±0.09 & 0.46±0.09 & 0.46±0.09 & 0.77±0.13 & 14169.84 
& 0.45±0.11 & 0.45±0.11 & 0.45±0.11 & 0.78±0.15 & 45118.80 \\
& CMB 
& 0.48±0.23 & 0.48±0.23 & 0.48±0.23 & 0.73±0.33 & 7933.74  
& 0.42±0.19 & 0.42±0.19 & 0.42±0.19 & 0.82±0.26 & 11783.44 \\
& GraN-LCS 
& 0.39±0.14   & 0.39±0.14   & 0.39±0.14   & 0.86±0.20   & -         
& 0.43±0.13   & 0.43±0.13   & 0.43±0.13   & 0.80±0.18   & -   \\
&MMB-by-MMB 
& \textbf{0.92±0.18} & \textbf{0.92±0.18} & \textbf{0.92±0.18} & \textbf{0.11±0.25} & \textbf{1054.52} 
& \textbf{0.92±0.20} & \textbf{0.92±0.20} & \textbf{0.92±0.20} & \textbf{0.11±0.29} & \textbf{2029.79}   \\
\hline
& PC-stable 
& 0.72±0.08 & 0.72±0.08 & 0.72±0.08 & 0.40±0.12 & 12629.93 
& 0.73±0.07 & 0.73±0.07 & 0.73±0.07 & 0.39±0.10 & 25137.35  \\
& FCI 
& 0.97±0.10 & 0.97±0.12 & 0.97±0.11 & 0.05±0.15 & 25540.04
& 0.97±0.11 & 0.97±0.12 & 0.97±0.11 & 0.04±0.16 & 42790.89\\
& RFCI 
& 0.94±0.15 & 0.93±0.17 & 0.94±0.16 & 0.09±0.23 & 12629.93 
& 0.97±0.11 & 0.96±0.12 & 0.96±0.12 & 0.05±0.17 & 25137.35 \\
{Problem1}& MB-by-MB 
& 0.72±0.22 & 0.78±0.24 & 0.73±0.22 & 0.37±0.31 & 16082.27 
&NA &NA &NA &NA &NA \\
& CMB 
& 0.62±0.13 & 0.63±0.14 & 0.62±0.13 & 0.54±0.19 & 11702.02 
& 0.63±0.13 & 0.63±0.13 & 0.63±0.13 & 0.53±0.18 & 103663.78 \\
& GraN-LCS 
& 0.57±0.15 & 0.57±0.16 & 0.57±0.15 & 0.61±0.21 & - 
& 0.65±0.21 & 0.66±0.22 & 0.65±0.21 & 0.49±0.30 & - \\
&MMB-by-MMB 
& \textbf{0.97±0.11} & \textbf{0.97±0.14} & \textbf{0.97±0.13} & \textbf{0.05±0.19} & \textbf{2895.81} 
& \textbf{0.99±0.06} & \textbf{0.99±0.07} & \textbf{0.99±0.07} & \textbf{0.02±0.10} & \textbf{6931.07}  \\
\hline
& PC-stable 
& 0.23±0.06   & 0.23±0.06   & 0.23±0.06   & 1.08±0.08   & 12715.64     
& 0.23±0.06   & 0.23±0.06   & 0.23±0.06   & 1.08±0.08   & 26274.96  \\
& FCI 
& 0.79±0.16   & 0.70±0.19   & 0.73±0.18   & 0.39±0.25   & 25501.34   
& 0.78±0.21   & 0.73±0.22   & 0.74±0.22   & 0.37±0.31   & 43773.13  \\
& RFCI 
& 0.71±0.19   & 0.62±0.20   & 0.65±0.19   & 0.50±0.27   & 12715.64   
& 0.75±0.23   & 0.70±0.24   & 0.71±0.23   & 0.41±0.33   & 26274.96   \\
{GDIOUT}& MB-by-MB 
& 0.31±0.12 & 0.44±0.18 & 0.34±0.13 & 0.92±0.19 & 14327.88
&NA &NA &NA &NA &NA \\
& CMB 
& 0.27±0.09   & 0.37±0.15   & 0.29±0.10   & 0.99±0.15   & 6279.86    
& 0.26±0.07   & 0.33±0.14   & 0.27±0.08   & 1.02±0.13   & \textbf{6850.83} \\
& GraN-LCS 
& 0.24±0.08   & 0.26±0.11   & 0.25±0.08   & 1.06±0.12   & -        
& 0.25±0.10   & 0.28±0.14   & 0.26±0.10   & 1.05±0.15   & -     \\
&MMB-by-MMB 
& \textbf{0.87±0.17} & \textbf{0.87±0.19} & \textbf{0.85±0.18} & \textbf{0.22±0.26} & \textbf{5514.87}     
& \textbf{0.87±0.25} & \textbf{0.87±0.26} & \textbf{0.86±0.25} & \textbf{0.20±0.36} & \underline{14635.54}  \\

\hline
& PC-stable 
& 0.24±0.10 & 0.24±0.10 & 0.24±0.10 & 1.07±0.14 & 12566.90    
& 0.24±0.10 & 0.24±0.10 & 0.24±0.10 & 1.07±0.14 & 26315.51  \\
& FCI 
& 0.80±0.18 & 0.68±0.17 & 0.71±0.17 & 0.40±0.24 & 25398.65    
& 0.77±0.12 & 0.71±0.14 & 0.73±0.13 & 0.39±0.18 & 44028.39  \\
& RFCI 
& 0.72±0.22 & 0.61±0.20 & 0.64±0.20 & 0.51±0.29 & 12566.90     
& 0.76±0.13 & 0.69±0.14 & 0.71±0.13 & 0.41±0.19 & 26315.51  \\
{PrData}& MB-by-MB 
& 0.32±0.14 & 0.47±0.20 & 0.35±0.15 & 0.90±0.22 & 69692.41 
&NA &NA &NA &NA &NA \\
& CMB 
& 0.30±0.15 & 0.39±0.19 & 0.31±0.15 & 0.96±0.22 & 11953.29
& 0.27±0.11 & 0.37±0.17 & 0.28±0.11 & 0.99±0.17 & 69763.20 \\
& GraN-LCS 
& 0.24±0.08 & 0.25±0.09 & 0.24±0.08 & 1.07±0.11 & - 
& 0.25±0.10 & 0.26±0.11 & 0.25±0.10 & 1.06±0.14 & - \\
&MMB-by-MMB 
& \textbf{0.84±0.15} & \textbf{0.77±0.15} & \textbf{0.77±0.14} & \textbf{0.32±0.19} & \textbf{6833.29}     
& \textbf{0.90±0.13} & \textbf{0.90±0.14} & \textbf{0.88±0.14} & \textbf{0.18±0.19} & \textbf{15581.97}  \\

\bottomrule
\end{tabular}
}
\label{table-7}
\begin{tablenotes}
		\item \scriptsize{Note: The symbol '-' indicates that GraN-LCS does not output this information. MMB-by-MMB with the second best result is underlined. $\uparrow$ means a higher value is better, and vice versa. NA entries for MB-by-MB demonstrate that the runtime exceeds a certain threshold.}
\end{tablenotes}
\end{table*}


\begin{table*}[hpt!]
\centering
\small
\caption{Performance Comparisons on ANDES.Net}
\resizebox{0.98\textwidth}{!}{
\begin{tabular}{cc|ccccc|ccccc}
\toprule
\multicolumn{2}{c}{} & \multicolumn{5}{|c|}{Size=1000} & \multicolumn{5}{c}{Size=5000}\\
\hline
\multicolumn{1}{c}{Target} & Algorithm & {\textbf{Precision}$\uparrow$} & {\textbf{Recall}$\uparrow$} &{\textbf{F1}$\uparrow$} & {\textbf{Distance}$\downarrow$} & {\textbf{nTest}$\downarrow$} & {\textbf{Precision}}$\uparrow$ & {\textbf{Recall}$\uparrow$} &{\textbf{F1}$\uparrow$} & {\textbf{Distance}$\downarrow$} & {\textbf{nTest}$\downarrow$}\\
\hline
& PC-stable 
&0.23±0.07 & 0.23±0.07 & 0.23±0.07 & 1.09±0.10 & 234677.37 
& 0.23±0.07 & 0.23±0.07 & 0.23±0.07 & 1.09±0.10 & 439483.08  \\
& FCI 
& 0.70±0.24 & 0.68±0.25 & 0.68±0.24 & 0.45±0.34 & 901063.34 
& 0.79±0.24 & 0.78±0.25 & 0.79±0.24 & 0.30±0.34 & 1584682.77  \\
& RFCI 
& 0.66±0.24 & 0.64±0.24 & 0.65±0.24 & 0.50±0.34 & 234677.37  
& 0.78±0.24 & 0.77±0.25 & 0.77±0.25 & 0.32±0.35 & 439483.08   \\
{RApp3($V_5$)} & MB-by-MB 
& 0.34±0.07 & 0.43±0.12 & 0.36±0.08 & 0.89±0.12 & 24239.83 
& 0.39±0.08 & 0.56±0.12 & 0.43±0.09 & 0.78±0.12 & 44225.00\\
& CMB 
& 0.33±0.06 & 0.38±0.09 & 0.34±0.07 & 0.92±0.09 & 79932.47 
& 0.32±0.05 & 0.39±0.09 & 0.34±0.07 & 0.93±0.09 & 145631.64 \\
& GraN-LCS 
& 0.39±0.12 & 0.46±0.16 & 0.40±0.12 & 0.84±0.17 & -           
& 0.38±0.13 & 0.42±0.16 & 0.39±0.13 & 0.86±0.19 & -        \\
&MMB-by-MMB 
& \textbf{0.91±0.15} & \textbf{0.90±0.17} & \textbf{0.89±0.17} & \textbf{0.16±0.24} & \textbf{5043.44}
& \textbf{0.98±0.07} & \textbf{0.98±0.07} & \textbf{0.98±0.07} & \textbf{0.03±0.10} & \textbf{4595.15}  \\
\hline
& PC-stable
& 0.26±0.04 & 0.26±0.04 & 0.26±0.04 & 1.05±0.06 & 234677.37 
& 0.25±0.00 & 0.25±0.00 & 0.25±0.00 & 1.06±0.00 & 439483.08 \\
& FCI 
& 0.87±0.16 & 0.80±0.21 & 0.82±0.19 & 0.25±0.27 & 901063.34 
& 0.90±0.16 & 0.87±0.19 & 0.88±0.18 & 0.17±0.25 & 1584682.77 \\
& RFCI 
& 0.85±0.18 & 0.79±0.22 & 0.81±0.21 & 0.28±0.29 & 234677.37  
& 0.88±0.17 & 0.85±0.20 & 0.86±0.19 & 0.20±0.27 & 439483.08   \\
{SNode-27} & MB-by-MB 
& 0.57±0.08 & 0.63±0.12 & 0.58±0.08 & 0.58±0.12 & 34010.75
&NA &NA &NA &NA &NA \\
& CMB 
& 0.56±0.06 & 0.61±0.10 & 0.58±0.08 & 0.60±0.11 & 126675.88 
& 0.56±0.07 & 0.61±0.09 & 0.57±0.07 & 0.60±0.10 & 257776.30 \\
& GraN-LCS 
& 0.67±0.14 & 0.70±0.16 & 0.66±0.13 & 0.47±0.19 & -       
& 0.71±0.15 & 0.71±0.16 & 0.69±0.14 & 0.44±0.20 & -       \\
&MMB-by-MMB 
& \textbf{0.95±0.15} & \textbf{0.93±0.18} & \textbf{0.94±0.17} & \textbf{0.09±0.24} & \textbf{4794.07}     
& \textbf{0.95±0.16} & \textbf{0.95±0.17} & \textbf{0.95±0.17} & \textbf{0.07±0.24} & \textbf{7104.74} \\
\hline
& PC-stable 
& 0.25±0.00 & 0.25±0.00 & 0.25±0.00 & 1.06±0.00 & 234677.37 
& 0.25±0.00 & 0.25±0.00 & 0.25±0.00 & 1.06±0.00 & 439483.08  \\
& FCI & 0.83±0.22 & 0.84±0.21 & 0.83±0.22 & 0.24±0.31 & 901063.34 
& 0.87±0.22 & 0.87±0.22 & 0.87±0.22 & 0.19±0.31 & 1584682.77 \\
& RFCI & 0.80±0.24 & 0.81±0.24 & 0.80±0.24 & 0.28±0.34 & 234677.37  
& 0.87±0.23 & 0.86±0.23 & 0.86±0.23 & 0.19±0.32 & 439483.08  \\
{SNode-21} & MB-by-MB 
& 0.31±0.12 & 0.42±0.20 & 0.33±0.14 & 0.93±0.20 & 18424.49  
& 0.33±0.10 & 0.49±0.13 & 0.36±0.11 & 0.88±0.15 & 82493.83  \\
&CMB      
& 0.31±0.14 & 0.36±0.17 & 0.32±0.14 & 0.96±0.20 & 94615.12 
& 0.30±0.13 & 0.35±0.16 & 0.31±0.13 & 0.97±0.19 & 232544.08\\
& GraN-LCS 
& 0.47±0.16 & 0.57±0.19 & 0.50±0.17 & 0.71±0.24 & -      
& 0.47±0.18 & 0.56±0.21 & 0.50±0.18 & 0.71±0.26 & -        \\
&MMB-by-MMB & \textbf{0.90±0.19} & \textbf{0.92±0.16} & \textbf{0.90±0.18} & \textbf{0.14±0.26} & \textbf{5086.75}    
& \textbf{0.96±0.10} & \textbf{0.97±0.09} & \textbf{0.96±0.10} & \textbf{0.05±0.14} & \textbf{7601.29}    \\
\hline
& PC-stable 
& 0.46±0.09 & 0.46±0.09 & 0.46±0.09 & 0.77±0.13 & 234677.37 
& 0.46±0.09 & 0.46±0.09 & 0.46±0.09 & 0.77±0.13 & 439483.08  \\
& FCI 
& 0.79±0.24 & 0.78±0.24 & 0.78±0.24 & 0.32±0.34 & 901063.34 
& 0.84±0.23 & 0.84±0.23 & 0.84±0.24 & 0.23±0.33 & 1584682.77  \\
& RFCI 
& 0.79±0.24 & 0.79±0.24 & 0.79±0.24 & 0.32±0.34 & 234677.37 
& 0.84±0.23 & 0.83±0.23 & 0.83±0.24 & 0.24±0.33 & 439483.08  \\
{RApp4} 
& MB-by-MB 
& 0.26±0.14 & 0.29±0.16 & 0.26±0.14 & 1.04±0.20 & 18761.00
&NA &NA &NA &NA &NA \\
& CMB 
& 0.25±0.16 & 0.26±0.17 & 0.25±0.16 & 1.06±0.23 & 87956.70
& 0.23±0.12 & 0.23±0.12 & 0.23±0.12 & 1.09±0.17 & 212387.71 \\
& GraN-LCS 
& 0.36±0.16 & 0.45±0.19 & 0.38±0.16 & 0.87±0.23 & -        
& 0.38±0.17 & 0.48±0.22 & 0.40±0.18 & 0.84±0.26 & -       \\
&MMB-by-MMB 
& \textbf{0.91±0.22} & \textbf{0.93±0.19} & \textbf{0.91±0.21} & \textbf{0.12±0.30} & \textbf{3153.11}     
& \textbf{0.97±0.12} & \textbf{0.98±0.09} & \textbf{0.97±0.11} & \textbf{0.04±0.15} & \textbf{1430.12}  \\
\hline
& Pc-stable        
& 0.46±0.09 & 0.46±0.09 & 0.46±0.09 & 0.76±0.13 & 234677.37
& 0.46±0.09 & 0.46±0.09 & 0.46±0.09 & 0.76±0.13 & 439483.08  \\
& FCI             
& 0.89±0.17 & 0.88±0.18 & 0.88±0.18 & 0.17±0.26 & 901063.34 
& 0.93±0.15 & 0.91±0.18 & 0.91±0.17 & 0.12±0.24 & 1584682.77 \\
& RFCI            
& 0.87±0.20 & 0.86±0.20 & 0.86±0.20 & 0.19±0.29 & 234677.37 
& 0.92±0.16 & 0.91±0.18 & 0.91±0.17 & 0.13±0.25 & 439483.08  \\
{SNode-4} & MB-by-MB 
0.23±0.12 & 0.24±0.13 & 0.23±0.12 & 1.08±0.18 & 21378.94
& 0.27±0.03 & 0.31±0.10 & 0.28±0.05 & 1.02±0.08 & 55156.33\\
& CMB 
& 0.25±0.16 & 0.26±0.17 & 0.26±0.16 & 1.05±0.23 & 120151.52 
& 0.24±0.17 & 0.25±0.17 & 0.24±0.16 & 1.07±0.23 & 248697.33 \\
& GraN-LCS 
& 0.35±0.12 & 0.47±0.14 & 0.37±0.12 & 0.87±0.17 & -         
& 0.37±0.15 & 0.45±0.18 & 0.38±0.15 & 0.86±0.22 & -  \\
&MMB-by-MMB 
& \textbf{0.90±0.22} & \textbf{0.92±0.17} & \textbf{0.90±0.21} & \textbf{0.14±0.29} & \textbf{3125.87}    
& \textbf{0.98±0.07} & \textbf{0.98±0.07} & \textbf{0.98±0.07} & \textbf{0.03±0.11} & \textbf{1913.94}      \\
\hline
& PC-stable 
& 0.47±0.08 & 0.47±0.08 & 0.47±0.08 & 0.75±0.12 & 234677.37 & 0.47±0.08 & 0.47±0.08 & 0.47±0.08 & 0.75±0.12 & 439483.08  \\
& FCI            
& 0.91±0.15 & 0.89±0.17 & 0.90±0.16 & 0.15±0.24 & 901063.34 
& 0.94±0.12 & 0.92±0.16 & 0.93±0.14 & 0.10±0.21 & 1584682.77 \\
& RFCI            
& 0.91±0.15 & 0.89±0.17 & 0.90±0.16 & 0.15±0.24 & 234677.37  
& 0.94±0.12 & 0.92±0.16 & 0.93±0.14 & 0.10±0.21 & 439483.08   \\
{SNode-47} & MB-by-MB 
& 0.27±0.12 & 0.30±0.14 & 0.28±0.12 & 1.01±0.18 & 14689.66   
& 0.23±0.13 & 0.26±0.14 & 0.24±0.13 & 1.07±0.18 & 55457.00  \\
& CMB      
& 0.26±0.13 & 0.26±0.13 & 0.26±0.13 & 1.04±0.19 & 124314.84
& 0.25±0.12 & 0.25±0.13 & 0.25±0.12 & 1.06±0.17 & 247077.40 \\
& GraN-LCS 
&0.32±0.13 & 0.41±0.18 & 0.34±0.14 & 0.92±0.20 & - 
&0.32±0.15 & 0.38±0.21 & 0.33±0.16 & 0.94±0.24 & -\\
&MMB-by-MMB 
& \textbf{0.94±0.18} & \textbf{0.95±0.13} & \textbf{0.94±0.16} & \textbf{0.08±0.23} & \textbf{1693.23}    
& \textbf{1.00±0.00} & \textbf{1.00±0.00} & \textbf{1.00±0.00} & \textbf{0.00±0.00} & \textbf{1590.80}    \\
\hline
& PC-stable 
& 0.47±0.08 & 0.47±0.08 & 0.47±0.08 & 0.75±0.11 & 234677.37  
& 0.47±0.08 & 0.47±0.08 & 0.47±0.08 & 0.75±0.12 & 439483.08   \\
&FCI             
& 0.65±0.17 & 0.70±0.16 & 0.66±0.16 & 0.48±0.23 & 901063.34
& 0.72±0.19 & 0.79±0.15 & 0.74±0.17 & 0.37±0.25 & 1584682.77 \\
&RFCI           
& 0.63±0.18 & 0.68±0.17 & 0.65±0.17 & 0.50±0.24 & 234677.37 
& 0.71±0.19 & 0.79±0.16 & 0.74±0.18 & 0.37±0.25 & 439483.08   \\
{SNode-24} & MB-by-MB 
& 0.40±0.16 & 0.48±0.23 & 0.42±0.17 & 0.82±0.25 & 20934.91
&NA &NA &NA &NA &NA \\
&CMB      
& 0.52±0.11 & 0.58±0.17 & 0.54±0.13 & 0.65±0.18 & 54165.45
& 0.54±0.09 & 0.61±0.15 & 0.56±0.10 & 0.62±0.14 & 69412.77\\
&GraN-LCS 
& 0.51±0.12 & 0.60±0.13 & 0.53±0.12 & 0.66±0.17 & -      
& 0.56±0.11 & 0.63±0.15 & 0.58±0.12 & 0.60±0.17 & -       \\
&MMB-by-MMB 
& \textbf{0.88±0.20} & \textbf{0.90±0.18} & \textbf{0.88±0.20} & \textbf{0.17±0.28} & \textbf{7202.50}    
& \textbf{0.99±0.06} & \textbf{0.99±0.06} & \textbf{0.99±0.07} & \textbf{0.02±0.09} & \textbf{15345.62 }  \\
\bottomrule
\end{tabular}
}
\label{table-8}
\begin{tablenotes}
		\item \scriptsize{Note: The symbol '-' indicates that GraN-LCS does not output this information. MMB-by-MMB with the second best result is underlined. $\uparrow$ means a higher value is better, and vice versa. NA entries for MB-by-MB demonstrate that the runtime exceeds a certain threshold.}
\end{tablenotes}
\end{table*}


\end{document}